\documentclass{article}

\PassOptionsToPackage{numbers, compress}{natbib}

\usepackage[preprint]{neurips_2025}

\usepackage[utf8]{inputenc} 
\usepackage[T1]{fontenc}    
\usepackage{hyperref}       
\usepackage{url}            
\usepackage{booktabs}       
\usepackage{amsfonts}       
\usepackage{nicefrac}       
\usepackage{microtype}      
\usepackage{xcolor}         

\usepackage{graphicx}
\usepackage{subfigure}

\usepackage{colortbl}
\usepackage{multirow}
\newcommand{\benchtitle}{$\mathcal{V}imo{RAG}$\xspace}

\title{\benchtitle: Video-based Retrieval-augmented\\ 3D Motion Generation for Motion Language Models}

\usepackage[framemethod=TikZ]{mdframed}
\newcommand{\gray}[1]{\textcolor[rgb]{0.6,0.6,0.6}{#1}}
\usepackage{pgf-pie}  
\definecolor{c1}{HTML}{8AC6D1}
\definecolor{c2}{HTML}{BBDED6}
\definecolor{c3}{HTML}{FAE3D9}
\definecolor{c4}{HTML}{FFB6B9}
\usepackage{pgfplots}
\usepackage{tikz}
\usepackage{svg}
\usetikzlibrary{patterns}
\definecolor{RYB1}{RGB}{255,151,151}
\definecolor{RYB2}{RGB}{255,188,113}
\definecolor{RYB3}{RGB}{150,204,236}
\definecolor{RYB4}{RGB}{119,190,182}

\usepackage{amsmath}
\usepackage{amssymb}
\usepackage{mathtools}
\usepackage{amsthm}
\usepackage{xspace}
\usepackage[capitalize,noabbrev]{cleveref}

\usepackage{paralist}
\usepackage[textsize=tiny]{todonotes}
\usepackage{textcomp}
\usepackage{wrapfig}
\usepackage{enumitem}

\author{%
  Haidong Xu$^1$, \, Guangwei Xu, \, Zhedong Zheng$^{2}$, \, Xiatian Zhu$^3$, \\ \textbf{ Wei Ji$^4$,\, Xiangtai Li$^5$,\, Ruijie Guo, \, Meishan Zhang$^1$, \, Min Zhang$^1$, \, Hao Fei$^6$\thanks{Corresponding Author: Hao Fei}}\\
  $^1$ Harbin Institute of Technology (Shenzhen)\, $^2$ University of Macau\, $^3$ University of Surrey\\ [2pt] $^4$ Nanjing University\,
   $^5$ Nanyang Technological University\, $^6$ National University of Singapore\\
  \texttt{182haidong@gmail.com, haofei7419@gmail.com}
}

\begin{document}

\maketitle

\vspace{-3mm}
\begin{abstract}
This paper introduces \textbf{\texttt{VimoRAG}}, a novel video-based retrieval-augmented motion generation framework for motion large language models (LLMs). 
As motion LLMs face severe out-of-domain/out-of-vocabulary issues due to limited annotated data, \textbf{\texttt{VimoRAG}} leverages large-scale in-the-wild video databases to enhance 3D motion generation by retrieving relevant 2D human motion signals. 
While video-based motion RAG is nontrivial, we address two key bottlenecks: 
(1) developing an effective motion-centered video retrieval model that distinguishes human poses and actions, 
and (2) mitigating the issue of error propagation caused by suboptimal retrieval results.
We design the Gemini Motion Video Retriever mechanism and the Motion-centric Dual-alignment DPO Trainer, enabling effective retrieval and generation processes. 
Experimental results show that \textbf{\texttt{VimoRAG}} significantly boosts the performance of motion LLMs constrained to text-only input. 
All the resources (\url{https://walkermitty.github.io/VimoRAG/}) are available.
\end{abstract}

\section{Introduction}
Generating diverse and realistic human motions from free-form text prompts has significant practical applications, including video gaming, robotic assistance, and virtual reality.
Previous advancements, ranging from transformer-based VAEs \cite{t2m-gpt} to recent diffusion-based generative models \cite{motion_diffuse}, have led to an increasingly promising generative performance.
With the emergence of LLMs, motion-language models (aka. motion LLMs) have been proposed \cite{motiongpt_aaai,motiongpt_nips}. 
These unified architectures not only understand various motions but also support motion generation.
To achieve competitive capabilities, motion LLMs require training on extremely large-scale datasets to ensure sufficient capacity.
Particularly for motion generation, a substantial amount of labeled data (i.e., text-motion pairs) is essential, without which, models face severe out-of-domain (OOD) and out-of-vocabulary (OOV) issues, making it difficult to generalize to the vast variety of dynamic human motions.

However, existing datasets of text-motion pairs are severely limited in scale, comprising only approximately 14k samples~\cite{humanml3d}, while the cost of large-scale annotation is prohibitively high.
To address this issue, \textit{Zhang et al.} introduce ReMoDiffuse \cite{remodiffuse}, a retrieval-augmented generation (RAG) method that retrieves relevant supplementary supervision signals from 3D motion databases.
While this method provides a promising direction to address the scarcity of labeled data, their effectiveness might still remain constrained by the size of existing 3D motion databases, i.e., totaling only 14k samples from datasets such as HumanML3D \cite{humanml3d}.

\begin{figure*}[!t]
    \centering
    \includegraphics[width=0.98\linewidth]{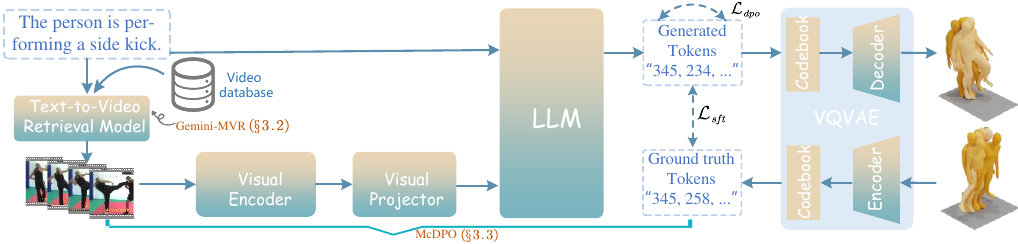}
    \vspace{-3mm}
    \caption{Overview of the VimoRAG pipeline: (1) text-to-video retrieval via Gemini-MVR, and (2) video-augmented motion generation guided by text and retrieved video. Gemini-MVR (Sec. \ref{Gemini-MVR}) is designed to improve cross-modal human-centric video retrieval, while the McDPO training strategy (Sec. \ref{ViMo}) mitigates error propagation caused by noisy retrievals.}
    \vspace{-5mm}
    \label{intro_overview}
\end{figure*}

To address this, this paper explores an innovative RAG paradigm: \emph{retrieving information from larger-scale in-the-wild videos to supplement abundant motion signals that can enhance motion generation}.
Although videos represent a 2D visual modality, intuitively, the 2D human motions depicted in videos inherently share similar characteristics with 3D human motions \cite{qiu2024vimo}, which can be utilized to guide the learning process of motion LLMs.
Most importantly, existing video data is highly accessible and virtually unlimited in scale, and in-the-wild videos capture diverse and unconstrained human motions, offering strong potential to address OOD/V challenges.
To this end, we introduce a simple but effective framework, named \textbf{\texttt{VimoRAG}} (cf. Fig.~\ref{intro_overview}).
VimoRAG first retrieves a video from an unlabeled video database based on the input text, then inputs both the text and the retrieved videos into an LLM to generate motion tokens, which are finally decoded into a motion sequence using VQ-VAE~\cite{vq-vae}.
\begin{wrapfigure}{r}{0.5\textwidth} 
    \centering
    \includegraphics[width=0.98\linewidth]{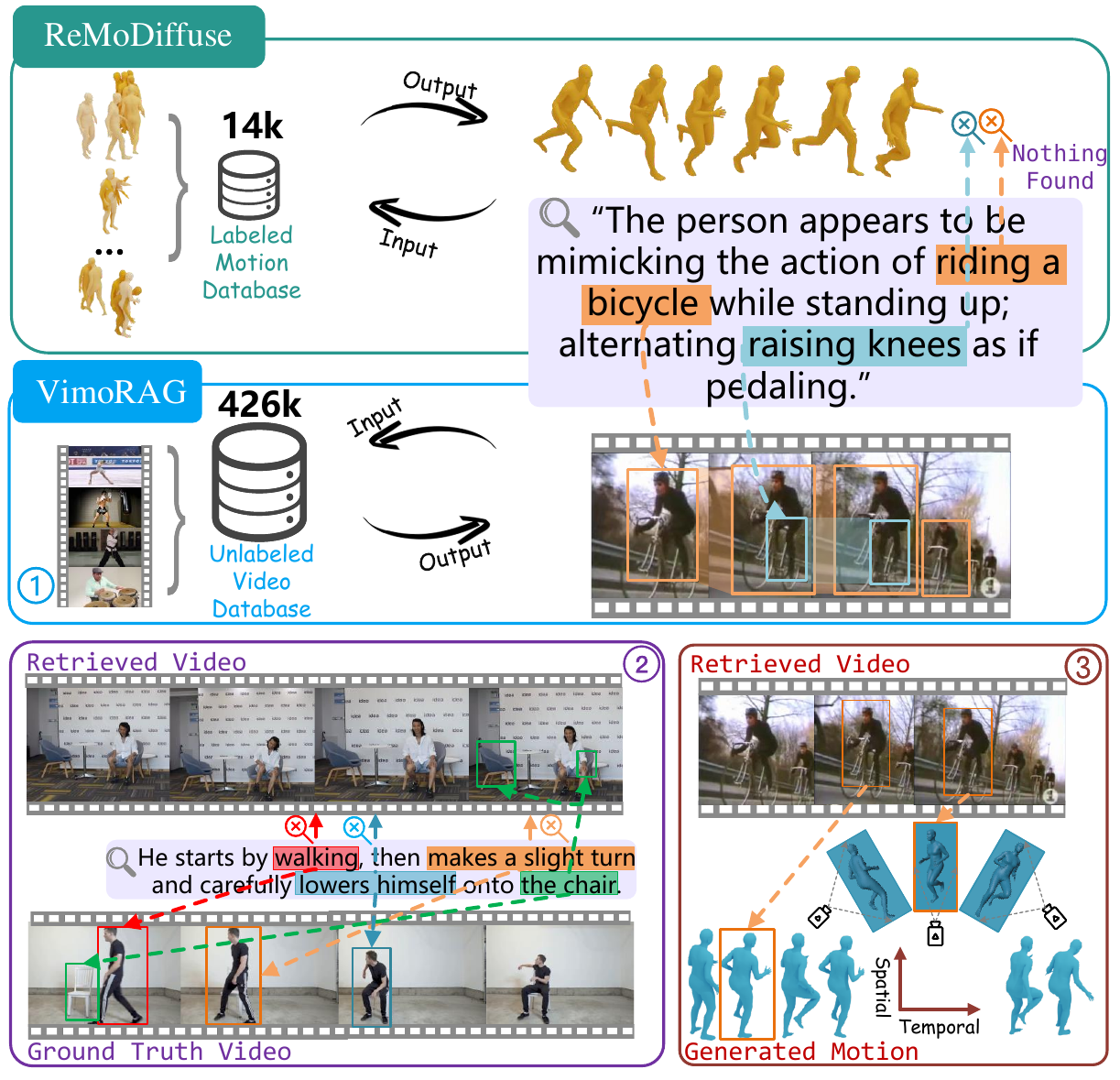}
    \caption{{ReMoDiffuse} is a RAG-based motion generation method, which is limited by the small scale of motion data and its reliance on annotated captions.
    We propose \textbf{\texttt{VimoRAG}}, which advances in 
    \textcircled{1} enabling retrieval from large-scale, \textbf{in-the-wild video databases without text captions}. \textcircled{2} Identifying and overcoming key challenges in human-centric text-to-video retrieval. 
    \textcircled{3} Ensuring alignment between retrieved videos and generated motions while mitigating error propagation.
    }
    \label{intro}
\vspace{-5mm}
\end{wrapfigure}

Yet this framework is nontrivial, facing at least two major challenges within such a new paradigm
\textbf{$\blacktriangleright$Challenge-I: \emph{Current video foundation models (VFMs) perform poorly in retrieving human-centric videos.}}
Our preliminary experiments indicate that existing VFMs, despite performing well on general-purpose video retrieval tasks (e.g., excelling at recognizing objects and attributes), struggle to distinguish human poses, actions, and behaviors in videos (cf. Fig.~\ref{intro}).
\textbf{$\blacktriangleright$Challenge-II: \emph{The error propagation caused by inaccurate or low-quality retrieved videos.}}
When the retrieval quality is poor, it significantly affects the quality of the generated content. Unfortunately, this issue is not thoroughly investigated in ReMoDiffuse~\cite{remodiffuse}.

To address the first challenge,
we design a Gemini Motion Video Retriever (termed \emph{Gemini-MVR}). 
Gemini-MVR incorporates dual fine-grained retrieval channels, at the action level and object level respectively, where a keypoints-aware router assigns weights to these two retrievers, allowing the system to simultaneously focus on human pose features and object information in complex videos, thereby improving the accuracy of human-centric video retrieval.

To address the second challenge, we propose a Motion-centric Dual-alignment DPO training strategy (named \emph{McDPO}). 
McDPO is to guide the LLM on how to utilize the prior information from the retrieval video (i.e., when to use it, when not to use it, and how much to rely on it) by performing self-correction.

We leverage a lightweight LLM Phi3-3.8B~\cite{phi-3} following \textit{Maaz et al.}~\cite{videogpt+} to evaluate the performance of this framework.
To evaluate its effectiveness, we conduct both cross-domain and in-domain experiments. 
To explore its potential, we conduct scaling experiments with varying retrieval corpus sizes.
Specifically, in OOD scenarios, we conduct zero-shot experiments on the challenging IDEA400~\cite{motionx} test set, where we verify that VimoRAG exhibits strong generalization capabilities.
We evaluate in-domain performance on the representative HumanML3D benchmark and observe that VimoRAG consistently improves most of the metrics compared to existing motion LLMs that operate with text-only input, and further exhibit sustained performance gains as the retrieval corpus expands.
In summary, our contributions are as follows:
\vspace{-2mm}

\begin{itemize}[leftmargin=14pt,topsep=0pt]
    \item To our knowledge, this paper is the first to propose a novel paradigm of \textbf{video-based 3D motion RAG}, which significantly alleviates the motion data scarcity bottleneck in existing methods.
    \item We present the \textbf{\texttt{VimoRAG}} framework, which integrates two plug-and-play modules—\emph{Gemini-MVR} retriever and \emph{McDPO} trainer—to address two key bottlenecks: human-centric video retrieval and error propagation in cross-modal motion RAG pipelines.
    \item Experimental results demonstrate that VimoRAG achieves substantial performance improvements in OOD settings and consistently enhances vanilla motion LLMs in in-domain scenarios. Furthermore, it exhibits clear potential for further gains as the retrieval corpus expands.
\end{itemize}

\section{Related Work}
Motion generation \cite{motion_survey} has long been a hot research topic in the related community, aiming to generate human-like 3D motion based on a given prompt, such as text, action, or incomplete motion.
Text-to-motion is among the most significant task settings and has attracted substantial research attention \cite{momask,remodiffuse,omg,motionclip,plan_posture,fan2024textual,oohmg}.
For instance, T2M-GPT \cite{t2m-gpt} explores a generative framework utilizing VQ-VAE \cite{vq-vae} and Transformer \cite{transformer} for motion generation.
MDM \cite{mdm} introduces a diffusion-based \cite{diffusion} generative model trained across multiple motion tasks.
MLD \cite{mld} enhances the latent diffusion model to produce motions conditioned on various inputs.
These motion generation specialists, following in-house training, deliver high performance in motion generation.

In recent years, the emergence of LLMs \cite{openagi} has demonstrated unprecedented levels of intelligence, driving the evolution from specialists to generalists.
In the motion domain, motion language models \cite{motiongpt_aaai,motiongpt_nips,motionllm,motiongpt2} have been proposed, where motion-aware encoders are connected to a central LLM, leading to motion generalists (also named motion LLMs) capable of perceiving various motions.
Further, motion generators (e.g., diffusion or VQ-VAE models) are integrated to achieve unified motion LLMs for both comprehension and generation \cite{motiongpt_aaai}.
To achieve robust motion manipulation capabilities, these motion LLMs require fine-tuning on large annotated datasets. 
However, compared to comprehension tasks, motion generation is more reliant on data, but motion annotation datasets are often quite limited due to the high cost of annotation.

In a recent study, Remodiffuse \cite{remodiffuse} introduces a motion generation method based on the retrieval-augmented generation (RAG) paradigm. It performs text-to-text retrieval from a labeled 3D motion database to fetch related motion signals and enhance generation quality. However, as previously mentioned, existing motion databases are typically limited in scale. In contrast, large-scale video databases are more accessible and diverse. Motivated by this, we explore a human motion-centric video retrieval framework to support 3D motion generation, where motion-consistent 2D features extracted from videos are effectively transferred to guide the 3D motion synthesis. Compared to Remodiffuse, our approach introduces two key innovations. First, we leverage cross-modal text-to-video retrieval to eliminate the reliance on motion databases that require manually written textual descriptions. Second, we are the first to identify the issue of error propagation in motion-RAG frameworks, and propose a novel algorithm, McDPO, to address it.

Notably, several motion LLM studies~\cite{motionbank,lmm,motionllm} have also explored human-related video tasks. Inspired by MotionBank, we construct our video corpus from multiple action-centric datasets. Unlike previous works that focus on building high-quality video collections, this work instead centers on validating the potential and robustness of VimoRAG framework when retrieving from a wild video corpus, and on addressing the potential inconsistency between video input and the generated motion.
\section{Our Approach: \texttt{VimoRAG}}
\label{method:vimorag}

\vspace{-1mm}

VimoRAG is a \textbf{Vi}deo-based \textbf{R}etrieval-\textbf{a}ugmented \textbf{Mo}tion \textbf{G}eneration framework.
We first introduce the overall architecture and our collected video database in Section \ref{preliminaries}.
Then we describe the details of two key components (Gimi-MVR and McDPO) in Section \ref{Gemini-MVR} and Section \ref{ViMo} respectively.

\vspace{-1mm}
\subsection{Preliminaries}
\vspace{-1mm}
\label{preliminaries}
\paragraph{Overall Architecture.} As depicted in Figure \ref{intro_overview}, VimoRAG is a pipeline composed of two essential steps. The initial step involves text-to-video retrieval, in which a motion description text is used to retrieve semantically relevant videos (the rank-1 video is used in this paper) from an unlabeled wild video database with our Gimi-MVR model. The subsequent step involves video-augmented motion generation, where both the text and retrieved videos are fed into the generation model to produce the motion sequence. 
Leveraging our novel McDPO trainer, we facilitate the contextually aligned motion generation process.

\label{database}
\paragraph{Human-centric Video Database.} In theory, it is possible to retrieve the most semantically relevant videos from all available short videos on the internet to overcome the challenges of open-vocabulary text descriptions in the industry. 
However, in academia, in order to efficiently verify the feasibility of our method and to advance this research direction, we gather and filter a vast human-centric video database (\textbf{HcVD}) consisting of 425,988 videos, sourced from \cite{motionx,aslan,hmdb51,kinetics-400,penn_action,ucf-101,nturgb}. 

To train a better retrieval model on human-centric videos, we leverage the widely used Qwen2-VL~\cite{qwen2-vl} model to synthesize textual descriptions following \textit{Zhao et al}~\cite{humanomni}.
We clarify that the synthetic captions are used solely to train the retriever, and are not involved in the motion RAG pipeline. 
VimoRAG is a ready-to-use framework that requires neither large-scale annotated videos nor text-to-text retrieval.
Additionally, we enhance the dataset quality by employing AlphaPose~\cite{alphapose} to filter out videos without human detection. 
More details are presented in the Appendix~\ref{extended_database}.

\begin{wrapfigure}{r}{0.5\textwidth}
    \includegraphics[width=0.98\linewidth]{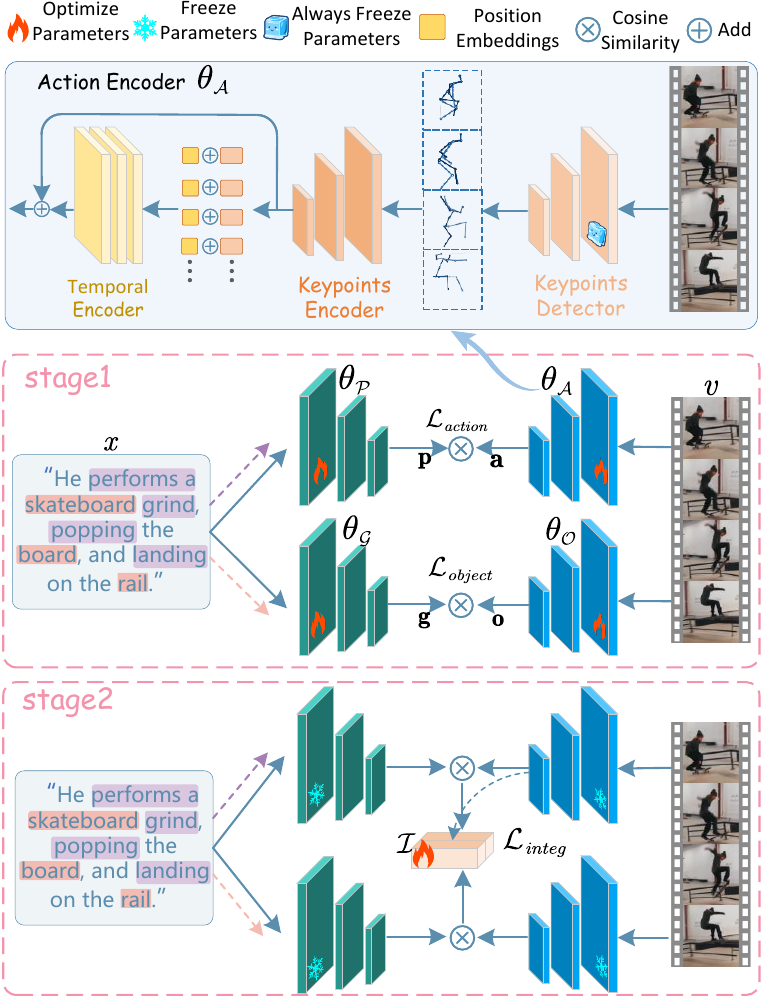}
        \vspace{-3mm}
    \caption{The architecture of the \textbf{Gemini-MVR} model. $\theta_{\mathcal{P}}$ and $\theta_{\mathcal{G}}$ represent the predicate semantic extractor and the argument semantic extractor, respectively. $\theta_{\mathcal{A}}$ and $\theta_{\mathcal{O}}$ denote the action encoder and object encoder, respectively. We simply introduce a lightweight action-level retriever and a routing module $\mathcal{I}$, while keeping the architecture of VFMs unchanged. This twin-module design provides strong extensibility.}
        \vspace{-5mm}
    \label{fig:method_Gemini-MVR}
\end{wrapfigure}


\subsection{Gemini Motion Video Retriever}
\label{Gemini-MVR}

\vspace{-1mm}

 Figure~\ref{fig:method_Gemini-MVR} illustrates the overall architecture of our method. The model builds upon the CLIP~\cite{clip} framework, where the text and video branches are encoded separately, and their similarity is computed for subsequent ranking.

Our model consists of two independent retrievers—object-level and action-level—whose outputs are fused by a lightweight router to produce the final similarity score. The object-level retriever captures visual entities (objects) and their textual arguments, while the action-level retriever targets motion and predicate-level semantics. 

\textbf{This design is driven by two main motivations}, as discussed in the introduction:
\textbf{First}, from our analysis of failure cases in existing VFMs, we observe that when a query describes only human actions without clear environmental or object cues, VFMs often struggle to retrieve the correct video.
To address this, we introduce a dedicated action-level retriever.
\textbf{Second}, we observe that many motions involve interaction with the environment, making object-level cues essential for understanding actions. While existing VFMs may struggle with behavior-only queries, they encode rich general knowledge from large-scale pretraining. To leverage this strength, we retain the VFM itself as the object-level retriever ($\theta_{\mathcal{G}}$ and $\theta_{\mathcal{O}}$ in Figure~\ref{fig:method_Gemini-MVR}).

Specifically,
for the action-level retriever, we carefully design an action encoder $\theta_\mathcal{A}$ and predicate semantic extractor $\theta_\mathcal{P}$.
In the action encoder, we first extract the 2D human keypoints from the $m$ frames of an input video $v$.
Then we encode each frame of keypoints to the feature space separately.
After this, we add learnable position embeddings frame by frame and feed them into a temporal encoder.
In order to avoid information loss in the sequential encoding process, the residual operation is adopted here.
To obtain a good initial weight,
we adopt the pretrained AlphaPose~\cite{alphapose} and the pretrained MotionBERT~\cite{motionbert} as the keypoints detector and encoder accordingly.
The temporal encoder is implemented by a transformer block~\cite{transformer} and is initialized randomly.
$\theta_\mathcal{P}$ is initialized from the text encoder in InternVideo~\cite{internvideo}.

We train the action-level retriever using the contrastive learning loss~\cite{clip4clip} $\mathcal{L}_{action} = \mathcal{L}_{p2a} + \mathcal{L}_{a2p}$
($\mathcal{L}_{a2p}$ is provided in the Appendix~\ref{extended_vimorag} due to space limitations):
\setlength\abovedisplayskip{3pt}
\setlength\belowdisplayskip{3pt}
\begin{equation}
    \mathcal{L}_{p2a} = - \frac{1}{B} \sum_{i}^{B} \log \frac{\exp(s(\textbf{p}_{i},\textbf{a}_{i}))}{ {\textstyle \sum_{j=1}^{B}\exp(s(\textbf{p}_{i},\textbf{a}_{j}))} } \,,    
\end{equation}

where $\mathbf{p}$ and $\mathbf{a}$ are embedding vectors encoded by $\theta_{\mathcal{P}}$ and $\theta_{\mathcal{A}}$ for the input value $t$ and $v$, respectively. 
$B$ denotes the number of text-video pairs in a training batch.
$s(,)$ denotes the cosine similarity.
For the object-level retriever, we directly adopt InternVideo~\cite{internvideo}, one of the most widely used VFMs, owing to its extensive common knowledge. 
The argument semantic extractor $\theta_{\mathcal{G}}$ and object encoder $\theta_{\mathcal{O}}$ are initialized using the text encoder and video encoder of InternVideo. 
During the fine-tuning stage, we utilize a loss function $\mathcal{L}_{object}$ similar to $\mathcal{L}_{action}$.
Let $\textbf{g}$ and $\textbf{o}$ denote the embedding vectors encoded by $\theta_{\mathcal{G}}$ and $\theta_{\mathcal{O}}$ for the input value $t$ and $v$, respectively.
We just replace $\textbf{p}$ with $\textbf{g}$ and replace $\textbf{a}$ with $\textbf{o}$ in $\mathcal{L}_{action}$, and then we can get the symmetric loss function $\mathcal{L}_{object}$.
It is worth emphasizing that the two semantic extractors do not explicitly extract the semantics of predicates and arguments.
We hypothesize that, through contrastive learning with different video features, each semantic extractor implicitly captures its respective focus—predicates and arguments.

Following the independent training of the two retrievers in stage 1, the subsequent step involves determining an effective approach to integrate the two retrievers.
A key consideration is that the allocation of weights should adapt to the characteristics of different motion videos. 
Building on this foundation, we propose an action-aware similarity integrator model, denoted as $\mathcal{I}$.
Considering the two-level retrieval models can be significantly large, the optimal $\mathcal{I}$ should be sufficiently lightweight to minimize retrieval delay.
To achieve this, we employ a simple linear method.
The cosine similarity $s(t,v)$ is calculated as follows:
\setlength\abovedisplayskip{3pt}
\setlength\belowdisplayskip{3pt}
\begin{equation}
\label{sim_equ}
    s(t ,v ) = \frac{\mathcal{I} _{0}(\mathbf{a} ) s(\mathbf{p} ,\mathbf{a} )}{\mathcal{I}_{0}(\mathbf{a} ) + \mathcal{I} _{1}(\mathbf{a} )}  + \frac{\mathcal{I} _{1}(\mathbf{a} ) s(\mathbf{g} ,\mathbf{o} )}{\mathcal{I} _{0}(\mathbf{a} ) + \mathcal{I} _{1}(\mathbf{a} )} \,,
\end{equation}
where $\mathcal{I} _{0}$ and $\mathcal{I} _{1}$ denote two output channels of $\mathcal{I}$.
In stage 2, the training loss function is $\mathcal{L}_{integ} = \mathcal{L}_{t2v} + \mathcal{L}_{v2t}$, where $\mathcal{L}_{t2v}$ is calculated as follows: 
\setlength\abovedisplayskip{3pt}
\setlength\belowdisplayskip{3pt}
\begin{equation}
    \mathcal{L}_{t2v} = - \frac{1}{B} \sum_{i}^{B} \log \frac{\exp(s(t_{i},v_{i}))}{ {\textstyle \sum_{j=1}^{B}\exp(s(t_{i},v_{j}))} }  \,.
\end{equation}
Notably, $\mathcal{L}_{v2t}$ and $\mathcal{L}_{t2v}$ exhibit a symmetric structure.
\begin{figure*}[!t]
    \centering
\includegraphics[width=.98\linewidth]{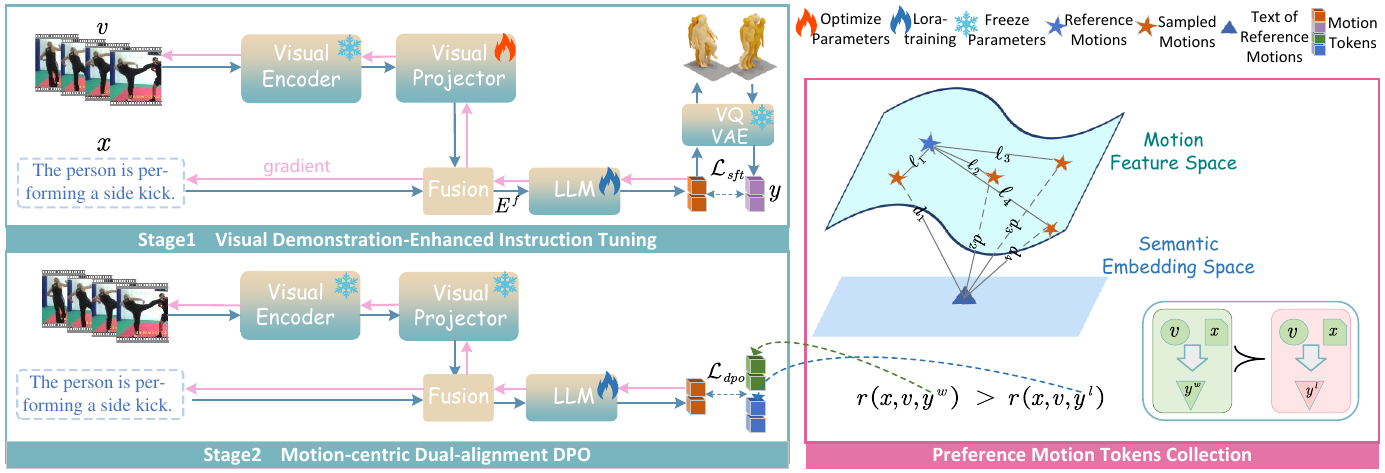}
    \caption{The \textbf{McDPO} training strategy. Given a text $t$ and a retrieved video $v$, we first perform visual demonstration-enhanced instruction tuning to establish a base reference model $\pi_{ref}$. Then, based on the motion-centric dual-alignment reward model, we construct a preference dataset and apply DPO training. The reward model jointly measures motion similarity in the feature space and semantic consistency with the text, guiding the model to learn informative motion priors and maximize preference rewards through self-improvement.}
    \label{fig::method_mcdpo}
\end{figure*}
\paragraph{Training and Inference.}
Since the action-level retriever is trained from scratch, while the object-level retriever has already been pretrained on a large corpus of text-video pairs, we first pretrain the action-level retriever using a subset of the HcVD
In Stage 1, we fine-tune both pretrained retrieval models in parallel, allowing for the optimization of all modules. 
In Stage 2, we freeze the two pretrained models and optimize only the similarity integrator. 
During the retrieval stage, we compute the similarity between the query text in the benchmark datasets for motion generation and each video in the HcVD using Equation \ref{sim_equ}.
More implementation details can be found in the Appendix~\ref{extended_vimorag}.

\subsection{Motion-centric Dual-alignment DPO Trainer}
\label{ViMo}
As illustrated in Figure~\ref{fig::method_mcdpo}, to fully leverage the descriptive information in the text and the rich 2D visual prior in the retrieved videos, we utilize LLM to project all the information from different modalities into the language space in Stage 1.
However, as mentioned in the introduction, there is an inherent gap between the 2D visual prior and the target 3D motion, as the motion prior in the retrieved videos represents only a sample of the full target motion space. 
Additionally, the 2D visual priors do not always align semantically with the text.
To guide the LLM to learn the appropriate direction for generation when such gaps arise, 
we construct the McDPO training set using a dual-alignment reward model, \textbf{allowing LLM to learn the most informative priors to maximize preference rewards by distinguishing its own struggle cases.}
We describe the details in the following parts.

\vspace{-3mm}
\paragraph{Visual Demonstration-Enhanced Instruction Tuning.}

Given an input text $x = \{x_{1}, x_{2},...x_{n_{x}}\}$ of $n_{x}$ tokens, a system prompt $\mathcal{S}$ of the LLM, an instruction template $\mathcal{T}$, and a retrieved video $v = \{ v_{1}, v_{2},...,v_{n_{v}}\}$ of $n_{v}$ frames, we first embed $v$ into $k$ segment-wise embeddings $E^v = \{ E^v_{1},E^v_{2},...,E^v_{k} \}$ following \textit{Maaz et al.}~\cite{videogpt+}
We then concatenate these elements to obtain the final input embeddings
$E^f = [emb(\mathcal{P}),\mathcal{T}(emb(x),E^v)]$
where $emb(\cdot)$ denotes the embedding layers of the LLM. 
For the target motion, we leverage the widely used VQ-VAE~\cite{vq-vae} to encode the continuous motion sequence into discrete motion tokens $y = \{y_{1},y_{2},..., y_{n_{y}}\}$ following MotionGPT~\cite{motiongpt_aaai}.
Inspired by them, we fine-tune the LLM using the following instruction-tuning format in Stage 1: 
\begin{mdframed}[backgroundcolor=gray!20,linecolor=white, roundcorner=5pt]
\textbf{System Prompt $\mathcal{P}$:} You are a helpful AI assistant. \\
\gray{\% $<|user|>$} \\
\textbf{Instruction Template $\mathcal{T}$:} Generate a sequence of motion tokens matching the following human motion description. You can use the video as a reference. Video information: \{ Retrieved Video $v$\} Motion description: \{Input Text $x$ \}\\
\gray{\% $<|assistant|>$} \\
\textbf{Answer $y$: } \{Sequence of Motion Tokens\}
\end{mdframed}

\vspace{-2mm}
The loss function $\mathcal{L}_{sft}$ in Stage 1 is as followed: 
$\mathcal{L} _{sft} = -  {\textstyle \sum_{n}^{} } \log p_{\theta}(y_{n}|y_{<n},E^f)$.

\begin{table*}[!t]
\fontsize{8.0}{8.0}\selectfont
\setlength{\tabcolsep}{2.0mm}
    \centering
    \renewcommand{\arraystretch}{0.8} 
    \caption{Zero-shot results on IDEA400 test set. All motions are generated by the models trained on HumanML3D training set. All results are reproduced using the officially released models (codes).
    VimoRAG achieves the best FID score, with other metrics closely matching SoTA.
    }
    \label{tab:idea400}
        \begin{tabular}{lcccccc}%
            \toprule
            \multirow{2}{*}{\textbf{Model}} &  \multirow{2}{*}{\textbf{FID} $\downarrow$}& \multicolumn{3}{c}{\textbf{R-Precision} $\uparrow$}  & \multirow{2}{*}{\textbf{MM Dist} $\downarrow$} & \multirow{2}{*}{\textbf{Diversity} $\uparrow$} \\
            \cmidrule{3-5}
            & & Top 1 &  Top 2 & Top 3 & \\
            \midrule
            \rowcolor{lightgray} \multicolumn{7}{l}{\bf\em $\bullet$ Motion Specialists}\\
            MoMask~\cite{momask}    & 5.982$^{\pm.089}$          & 0.110$^{\pm.003}$          & 0.195$^{\pm.006}$          & 0.266$^{\pm.006}$          & 5.625$^{\pm.023}$          & 7.558$^{\pm.119}$                           \\
            T2M-GPT~\cite{t2m-gpt}                   & 5.359$^{\pm.078}$ & 0.108$^{\pm.006}$ & 0.186$^{\pm.005}$ & 0.255$^{\pm.006}$          & 5.773$^{\pm.037}$ & 7.648$^{\pm.100}$                  \\
            MDM~\cite{mdm}                      & 5.907$^{\pm.107}$          & 0.113$^{\pm.004}$          & \textbf{0.200}$^{\pm.004}$          & \textbf{0.278}$^{\pm.004}$          & 6.013$^{\pm.020}$          & \textbf{8.131}$^{\pm.080}$            \\
            MotionDiffuse~\cite{motion_diffuse}                      & 5.485$^{\pm.038}$          & 0.110$^{\pm.002}$          & 0.194$^{\pm.002}$          & 0.266$^{\pm.003}$          & 6.038$^{\pm.005}$          & 6.884$^{\pm.095}$  \\
            MLD~\cite{mld}          & 5.410$^{\pm.085}$          & \textbf{0.114}$^{\pm.003}$          & 0.200$^{\pm.005}$          & 0.270$^{\pm.004}$ & 6.005$^{\pm.029}$          & 7.558$^{\pm.086}$           \\
            MotionGPT~\cite{motiongpt_nips}                              & 6.202$^{\pm.186}$ & 0.087$^{\pm.005}$ & 0.151$^{\pm.007}$ & 0.209$^{\pm.008}$ & 6.640$^{\pm.025}$ & 7.684$^{\pm.111}$             \\
            ReMoDiffuse~\cite{remodiffuse}          & 9.639$^{\pm.069}$          & 0.110$^{\pm.004}$          & 0.188$^{\pm.006}$          & 0.256$^{\pm.005}$ & \textbf{5.465}$^{\pm.015}$          & 7.540$^{\pm.120}$    \\
            \midrule
            \rowcolor{lightgray} \multicolumn{7}{l}{\bf\em $\bullet$ Motion LLMs}\\
            MotionGPT~\cite{motiongpt_aaai}        & 5.544$^{\pm.174}$          & 0.096$^{\pm.005}$          & 0.171$^{\pm.008}$          & 0.236$^{\pm.008}$          & 6.300$^{\pm.032}$          & 7.509$^{\pm.096}$   \\
            VimoRAG~(Ours)        & \textbf{2.388}$^{\pm.056}$          & 0.113$^{\pm.005}$          & 0.193$^{\pm.008}$          & 0.270$^{\pm.011}$          & 5.888$^{\pm.061}$          & 7.688$^{\pm.197}$ \\
            \bottomrule
        \end{tabular}%
\vspace{-3mm}
\end{table*}

\paragraph{Motion-centric Dual-alignment DPO.}

To empower the motion LLM with the ability to autonomously adapt to video priors of differing quality during the generation process, we introduce motion-centric
dual-alignment DPO training strategy.
As illustrated in Figure \ref{fig::method_mcdpo}, after the completion of Stage 1 training, we obtain a base reference model $\pi_{ref}$.
The videos used in Stage 1 are the retrieval results of the retrieval model, not ground truth, meaning that $\pi_{ref}$ has been learning to handle the potential gap between text and video during training.
However, $\pi_{ref}$ struggles to learn this aspect. 
In fact, we discover early in our experimens that $\pi_{ref}$’s performance on the training set is not stable, particularly when the gap is large. 
The experimental results shown in Figure~\ref{fig:mcdpo_study} support this observation (\texttt{NMC-R1} is worse than \texttt{MC-R1}).

To construct such a training set, we first use $\pi_{ref}$ to randomly sample $\kappa$ times to generate a motions' candidate set. 
In contrast to recent works~\cite{sheng2024explorin,pappa2024modipo}, which rely on human or AI-based proxies as reward models, we devise a more efficient dual-alignment reward model.
This model outputs a reward score for a sampled motion sequence $\hat{y}_{i}$ as follows:
\setlength\abovedisplayskip{3pt}
\setlength\belowdisplayskip{3pt}
\begin{equation}
\label{equ:reward}
r(x,v,\hat{y_{i}}) =-( w_{\ell} \frac{\ell(\hat{y_{i}},y)}{ {\textstyle \sum_{j\in\kappa }^{}}\ell(\hat{y}_{j},y) } + w_{d} \frac{d(\hat{y}_{i},x) }{\sum_{j\in\kappa}d(\hat{y}_{j},x)}) \,,
\end{equation}
where $\ell(\cdot)$ and $d(\cdot)$ 
denote the distribution distance and Euclidean distance computed based on the features of the two inputs, respectively.
$w_{\ell}$ and $w_{d}$ are hyper-parameters to respectively control the degree of alignment within the motion modality and between the text-motion modalities.
The reward model encourages the preferred motions to be closer to the reference motions in the feature space and more semantically aligned with the paired text in the semantic space.
By leveraging the reward model, we collect the chosen motions $y_{w}$ and the rejected motions $y_{l}$ from the existing motions' candidate set, thereby constructing a DPO dataset $\mathcal{D}_{dpo} = \{(x,v,y^{w},y^{l})\}$.
Finally, we adopt the following DPO training objective~\cite{dpo,llava-hound}.
Here $\pi_{\theta}$ and $\sigma$ denote the policy model and the logistic function, respectively. $\gamma$ here denotes the weighting coefficient.
\setlength\abovedisplayskip{3pt}
\setlength\belowdisplayskip{3pt}
\begin{equation}
\label{equ:dpo}
\begin{split}
    \mathcal{L}_{dpo} &= -\mathbb{E}_{(x,v,y^{w},y^{l})\sim \mathcal{D} _{dpo}} \Bigg[ \log \sigma \Big( \gamma \log \frac{\pi_{\theta}(y^{w}|x,v)}{\pi_{ref}(y^{w}|x,v)} 
    \quad - \gamma \log \frac{\pi_{\theta}(y^{l}|x,v)}{\pi_{ref}(y^{l}|x,v)} \Big) \Bigg] \,.
\end{split}
\end{equation}

\vspace{-5mm}
\paragraph{Training Strategy.}
We employ phi-3-mini~\cite{phi-3} as the backbone LLM and utilize the LoRA~\cite{lora} tuning method in both training stages. In Stage 1, we additionally tune the visual adapter while freezing the remaining modules. In Stage 2, all modules except for the LLM are kept frozen. Due to space constraints, further training configurations are detailed in the Appendix~\ref{extended_vimorag}.

\begin{table*}[!t]
\fontsize{8.0}{9.0}\selectfont
\setlength{\tabcolsep}{0.8mm}
    \centering
    \renewcommand{\arraystretch}{0.8} 
    \caption{Results on HumanML3D test set. ``*'' denotes results from original papers;
 others are reproduced using official code. \textbf{VimoRAG achieves the best FID and competitive performance across metrics among existing motion LLMs}. This framework significantly improves five metrics (highlighted in \textcolor{red}{red}) over MotionGPT~\cite{motiongpt_aaai} (Phi3-3.8B), \textbf{demonstrating the substantial advantage of incorporating video priors}. 
 The complete results with confidence intervals are shown in Table~\ref{tab:t2m_full}.}
    \label{tab:t2m}
    \begin{tabular}{llllllll}%
        \toprule
        \multirow{2}{*}{\textbf{Model}} & \multirow{2}{*}{\textbf{Backbone}}& \multirow{2}{*}{\textbf{FID} $\downarrow$}& \multicolumn{3}{c}{\textbf{R-Precision} $\uparrow$}  & \multirow{2}{*}{\textbf{MM Dist}$\downarrow$ } & \multirow{2}{*}{\textbf{Diversity}$\uparrow$ } \\
        \cmidrule{4-6}
        & & &Top 1 &  Top 2 & Top 3 & \\
        \midrule
        \rowcolor{lightgray} \multicolumn{8}{l}{\bf\em  Motion Specialists}\\
        MoMask~\cite{momask}    &--& 0.048          & 0.519          & 0.715          & 0.809          & 2.955          & 9.632                           \\
        T2M-GPT~\cite{t2m-gpt}    &--               & 0.112 & 0.489 & 0.679 & 0.774          & 3.125 & 9.691                     \\
        MDM~\cite{mdm}             &--         & 0.454          & 0.419          & 0.606          & 0.712          & 3.636          & 9.449            \\
        MotionDiffuse~\cite{motion_diffuse}  &                    --& 0.672          & 0.492          & 0.685          & 0.784          & 3.085          & 9.499  \\
        MLD~\cite{mld}   &       --& 0.425          & 0.468          & 0.656          & 0.759 & 3.266          & 9.698           \\
        ReMoDiffuse~\cite{remodiffuse}   &--       & 0.125          & 0.493          & 0.676          & 0.775 & 3.047          & 9.211    \\
        LMM*~\cite{lmm}      & --                       & 0.040 & 0.525 & 0.719 & 0.811 & 2.943 & 9.814             \\
        MotionLab*~\cite{motionlab} &--                             & 0.167 & -- & -- & 0.810 & 2.912 & 9.593             \\
        MotionLCM*~\cite{motionlcm} &--                             & 0.304 & 0.502 & 0.698 & 0.798 & 3.012 & 9.607             \\
        MotionCLR*~\cite{motionclr} &--                             & 0.269 & 0.544 & 0.732 & 0.831 & 2.806 & --             \\
        MotionGPT*~\cite{motiongpt_nips} &--                             & 0.232 & 0.492 & 0.681 & 0.778 & 3.096 & 9.528             \\
        BiPO*~\cite{bipo} &--                             & \textbf{0.030} & 0.523 & 0.714 & 0.809 & 2.880 & 9.556             \\
        StableMoFusion*~\cite{stable_mofusion} &--                             & 0.098 & 0.553 & 0.748 & 0.841 & -- & 9.748             \\
        MoGenTS*~\cite{mogents} &--                             & 0.033 & 0.529 & 0.719 & 0.812 & 2.867 & 9.570             \\
        LAMP*~\cite{lamp} &--                             & 0.032 & \textbf{0.557} & \textbf{0.751} & \textbf{0.843} & \textbf{2.759} & 9.571             \\
        \midrule
        \rowcolor{lightgray} \multicolumn{8}{l}{\bf\em  Motion LLMs}\\
        MotionGPT-2*~\cite{motiongpt2} &Llama3-8B                             & 0.191 & 0.496 & 0.691 & 0.782 & 3.080 & 9.860             \\
    
        MotionLLM*~\cite{motionagent} &GPT4+Gemma-2B                             & 0.230 & 0.515 & -- & 0.801 & 2.967 & \textbf{9.908}             \\
        \textit{Wang et al.}*~\cite{largemotionmodel} &Llama2-13B                             & 0.166 & 0.519 & -- & 0.803 & 2.964 & --             \\
        ScaMo*~\cite{scamo} &codesize 512-3B                             & 0.617 & 0.443 & 0.627 & 0.734 & 3.340 & 9.217             \\
        AvatarGPT*~\cite{avatargpt} &Llama-13B                             & 0.567 & 0.389 & 0.539 & 0.623 & -- & 9.489             \\
        MotionGPT*~\cite{motiongpt_aaai}  &Llama-13B      & 0.567          & --          & --          & --          & 3.775          & 9.006   \\
        \midrule
        MotionGPT~\cite{motiongpt_aaai}    & Phi3-3.8B    & 0.501          & 0.396          & 0.575          & 0.673          & 3.724          & 9.475 \\
        VimoRAG~(Ours)    & Phi3-3.8B    & $0.131\raisebox{-0.5ex}{\scriptsize\,\textcolor{red}{-73\%}}$

      & $0.452\raisebox{-0.5ex}{\scriptsize\,\textcolor{red}{+14\%}}$          & $0.655\raisebox{-0.5ex}{\scriptsize\,\textcolor{red}{+14\%}}$         & $0.764\raisebox{-0.5ex}{\scriptsize\,\textcolor{red}{+13\%}}$          &  $3.146\raisebox{-0.5ex}{\scriptsize\,\textcolor{red}{-15\%}}$        & $9.424\raisebox{-0.5ex}{\scriptsize\,-1\%}$\\
        \bottomrule
    \end{tabular}%
\end{table*}

\begin{figure*}
\vspace{1mm}
    \centering
    \includegraphics[width=0.99\linewidth]{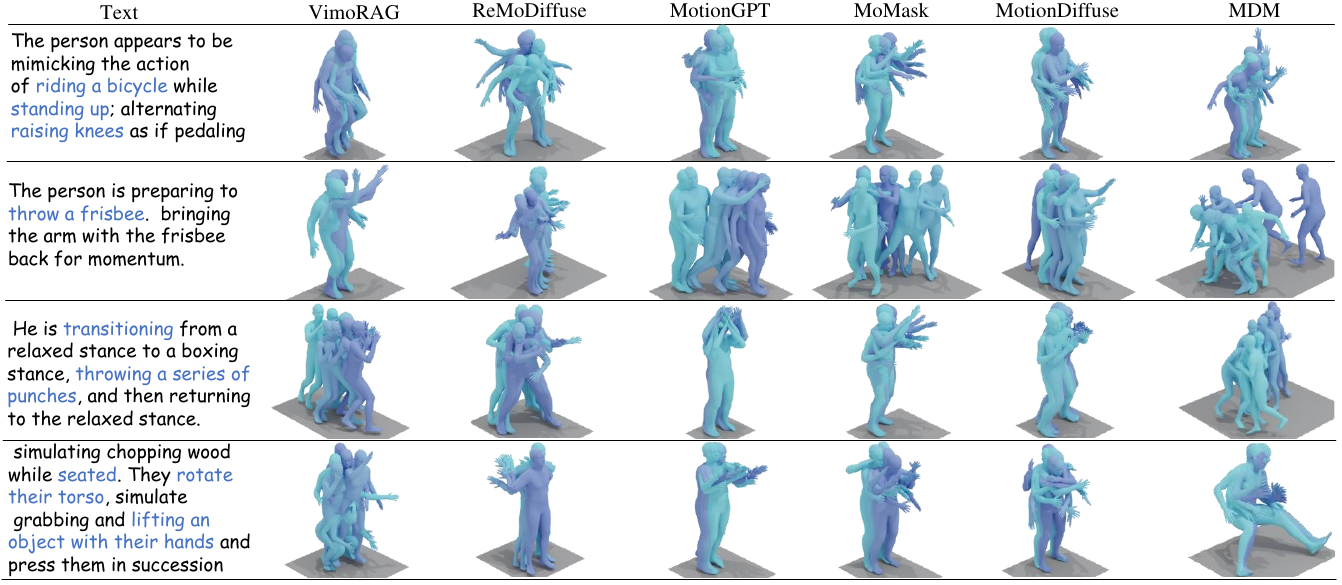}
        \vspace{-1mm}
    \caption{Zero-shot qualitative results on IDEA400 test set. All motions are directly generated by the models trained on HumanML3D training set. The text presented here only includes words related to motion due to space constraints. The full text and more results are available in Figure~\ref{fig:appendix1} and~\ref{fig:appendix2}.}
    \label{fig:idea400}
\end{figure*}

\vspace{-4mm}
\section{Experiment}

\vspace{-1mm}
\subsection{Datasets, Metrics and Baselines}
\vspace{-1mm}
We conduct extensive experiments on two widely used large-scale datasets following the existing works~\cite{plan_posture}. 
The first is the \textbf{IDEA400} dataset, a high-quality whole-body motion dataset composed of 12.5K clips and 2.6M frames in MotionX~\cite{motionx},
which is utilized to assess OOD performance. 
The other dataset, \textbf{HumanML3D}~\cite{humanml3d}, comprising 14,616 motion clips and 44,970 text descriptions, is utilized to evaluate in-domain performance.
For evaluation metrics, we adopt several widely recognized measures: \textit{Frechet Inception Distance} (FID), 
\textit{R-recall}, \textit{MultiModal Distance} (MM Dist), 
\textit{Diversity}.
More details can be found in Appendix~\ref{extended_experiment} due to the space limit.

\vspace{-2mm}
\subsection{Implementation Details}
\label{exper:details}
\vspace{-1mm}
We implement VimoRAG with PyTorch.
We use the same Gemini-MVR for all the text-to-motion experiments because the retrieval model is decoupled with the generation phrase in our framework.
In Stage 1 of McDPO, we train 2 epochs with a learning rate 2e-4 for LoRA parameters ($rank=128, \alpha=256$), with a learning rate 2e-5 for the visual adapter's parameters.
In Stage 2 of McDPO, we train 1 epoch with a learning rate 2e-4.
Inference is conducted using a single NVIDIA A800 GPU, while training is accelerated using 8 GPUs to enhance efficiency. Further details regarding the model configurations, training settings, and pose representation are provided in the Appendix~\ref{extended_experiment}.

\vspace{-2mm}
\subsection{Main Results}
\label{sec:main_results}
\vspace{-1mm}
\paragraph{Quantitative Results.}
Tables~\ref{tab:idea400} and \ref{tab:t2m} present a quantitative comparison between VimoRAG and SoTA techniques. 
For a fair comparison, each experiment is conducted ten times, and we report the results with a 95\% confidence interval. 
As illustrated in Table~\ref{tab:idea400}, VimoRAG achieves the best FID score, indicating its strong generalization capability in generating high-fidelity motions in OOD scenarios.
According to Table~\ref{tab:t2m}, VimoRAG \textbf{outperforms MotionGPT}~\cite{motiongpt_aaai} \textbf{by a large margin} across all metrics when using the same backbone. 
It achieves the best FID score and competitive performance on other metrics among motion LLMs based on the Phi-3 3.8B backbone. 
As shown in Figure \ref{fig:database_size1}, \textbf{performance improves steadily with larger retrieval sets, highlighting VimoRAG’s potential to enhance motion LLMs.}

\vspace{-3mm}
\paragraph{Qualitative Comparison.}
Figure~\ref{fig:idea400} demonstrates the qualitative comparison results on the IDEA400 test set. 
Among the results, those from VimoRAG appear to align more closely with the intended meaning of the given text.
The text descriptions in IDEA400 are quite intricate and differ significantly from those in the HumanML3D dataset. 
In the fourth demonstration, the text entails multiple changes in action, such as \textit{``seated, rotate, grabbing, lifting''}. 
Among all the motions showcased, the results from VimoRAG encompass more action types related to the text. 
Similar phenomena can be observed in other cases as well. 
More results are shown in Appendix (Section~\ref{extended_qualitative}) and our anonymous GitHub.

\vspace{-5mm}
\subsection{Ablation Study}
\label{ablation-study}

\begin{table}[t!]
\begin{minipage}{0.4\textwidth}  
    
    \fontsize{9}{9.5}\selectfont
    \setlength{\tabcolsep}{0.9mm}
    \raggedright
    \caption{Ablation study on HumanML3D validation set. \texttt{Gem} denotes Gemini-MVR retriever, \texttt{Mc} denotes McDPO, \texttt{Ran} denotes random retriever, \texttt{Int} denotes InternVideo retriever. The FID drop of \texttt{Gem+Mc} relative to each setting is highlighted in \color{blue}blue.}
    \begin{tabular}{llccc}
        \toprule
      
        \multirow{2}{*}{\textbf{Settings}} & \multirow{2}{*}{\textbf{FID}$\downarrow$}   & \multicolumn{3}{c}{\textbf{R-Precision}$\uparrow$}  \\ 
        \cmidrule{3-5}
        & & Top 1 & Top 2 & Top 3 \\
        \midrule
        Gem+Mc & 0.148 & 0.429 & 0.625 & 0.756 \\
        Ran+Mc & 0.544 {\color{blue}$_{\downarrow72.8\%}$} & 0.420 & 0.644 & 0.750 \\
        Int+Mc & 0.205 {\color{blue}$_{\downarrow27.8\%}$} & 0.433 & 0.638 & 0.736 \\
        Gem & 0.260 {\color{blue}$_{\downarrow43.1\%}$}& 0.403 & 0.582 & 0.682 \\
        \bottomrule
        \label{tab:ablation}
    \end{tabular}
    
\end{minipage}%
\hspace{0.05\textwidth}
\begin{minipage}{0.45\textwidth}  
\renewcommand{\arraystretch}{1.2} 
    \fontsize{9}{9.5}\selectfont
    \setlength{\tabcolsep}{1.2mm}
    \centering
        \caption{Text-to-video retrieval performance on HcVD test sets. Compared to the object-level VFM, Gemini-MVR achieves a significant improvement in the Recall@1 metric.}
 \begin{tabular}{llccc}
        \toprule

        \textbf{Retriever} & \textbf{R@1}$\uparrow$   & \textbf{R@5}$\uparrow$ & \textbf{R@10}$\uparrow$ & \textbf{MnR}$\downarrow$  \\ \midrule
                   \rowcolor{lightgray}
        \multicolumn{5}{c}{\bf\em Human-centric Video} \\
        InternVideo & 53.6 & 84.5 & 92.3 &  4.2 \\
        Gemini-MVR & 58.3 {\color{red}$\uparrow_{8.8\%}$}& 87.3 & 93.7 & 3.6 \\
        \midrule
                   \rowcolor{lightgray}
        \multicolumn{5}{c}{\bf\em Single Human-centric Video} \\
        InternVideo & 52.3 & 84.0 & 91.5 &  4.5 \\
        Gemini-MVR & 61.0 {\color{red}$\uparrow_{16.6\%}$} & 89.2 & 94.1 & 3.5 \\
        \bottomrule
        \label{tab:text-to-video}
    \end{tabular}
   
\end{minipage}
\vspace{-4mm}
\end{table}



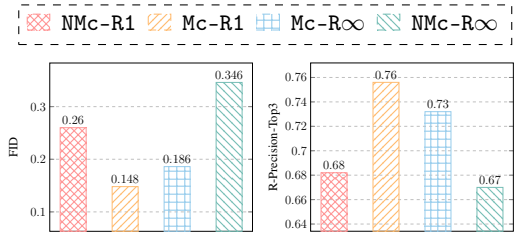
\begin{wrapfigure}{r}{0.5\textwidth}
    \centering
    \begin{tikzpicture}
        \begin{axis}[
            hide axis,
            height=2cm,
            xmin=0, xmax=1, ymin=0, ymax=1,
            legend columns=4,
            legend style={
                at={(0.2,0)},
                anchor=south west,
                font=\small,
                draw=black,
                dashed,
                column sep=0.5ex,
                legend image code/.code={
                    \draw[##1, bar width=3mm, yshift=-0.2em] plot coordinates {(0cm, 0.6em)};
                }
            },
            enlargelimits
        ]
            \addlegendimage{ybar,pattern=crosshatch,pattern color=RYB1,draw=RYB1} \addlegendentry{\texttt{NMc-R1}}
            \addlegendimage{ybar,pattern=north east lines,pattern color=RYB2,draw=RYB2} \addlegendentry{\texttt{Mc-R1}}
            \addlegendimage{ybar,pattern=grid,pattern color=RYB3,draw=RYB3} \addlegendentry{\texttt{Mc-R}$\infty$}
            \addlegendimage{ybar,pattern=north west lines,pattern color=RYB4,draw=RYB4} \addlegendentry{\texttt{NMc-R}$\infty$}
        \end{axis}
    \end{tikzpicture}

    \vspace{2mm}

    \resizebox{0.5\textwidth}{!}{
    \begin{tabular}{cc}
        \begin{tikzpicture}
            \begin{axis}[
                symbolic x coords={sftr1, mcr1, mcrn, sftrn},
                xtick=\empty,
                xticklabels=\empty,
                ylabel={FID},
                ylabel style={font=\large},
                yticklabel style={font=\large},
                ymajorgrids,
                bar width=25pt,
                ymin=0.10,
                grid style=dashed,
                nodes near coords,
                every node near coord/.append style={font=\large,/pgf/number format/precision=4,/pgf/number format/fixed},
                enlargelimits=0.15,
            ]
                \addplot[ybar,pattern=crosshatch,pattern color=RYB1,draw=RYB1] coordinates {(sftr1,0.260)};
                \addplot[ybar,pattern=north east lines,pattern color=RYB2,draw=RYB2] coordinates {(mcr1,0.148)};
                \addplot[ybar,pattern=grid,pattern color=RYB3,draw=RYB3] coordinates {(mcrn,0.186)};
                \addplot[ybar,pattern=north west lines,pattern color=RYB4,draw=RYB4] coordinates {(sftrn,0.346)};
            \end{axis}
        \end{tikzpicture}
        &
        \begin{tikzpicture}
            \begin{axis}[
                symbolic x coords={sftr1, mcr1, mcrn, sftrn},
                xtick=\empty,
                xticklabels=\empty,
                ylabel={R-Precision-Top3},
                ylabel style={font=\large},
                yticklabel style={font=\large},
                ymajorgrids,
                bar width=25pt,
                ymin=0.65,
                grid style=dashed,
                nodes near coords,
                every node near coord/.append style={font=\large},
                enlargelimits=0.15,
            ]
                \addplot[ybar,pattern=crosshatch,pattern color=RYB1,draw=RYB1] coordinates {(sftr1, 0.682)};
                \addplot[ybar,pattern=north east lines,pattern color=RYB2,draw=RYB2] coordinates {(mcr1, 0.756)};
                \addplot[ybar,pattern=grid,pattern color=RYB3,draw=RYB3] coordinates {(mcrn, 0.732)};
                \addplot[ybar,pattern=north west lines,pattern color=RYB4,draw=RYB4] coordinates {(sftrn, 0.670)};
            \end{axis}
        \end{tikzpicture}
    \end{tabular}
    }

    \caption{In-depth exploration of McDPO. \texttt{Mc} stands for McDPO setting, and \texttt{NMc} stands for No McDPO setting. \texttt{R1} indicates the use of rank-1 video during inference, while \texttt{R$\infty$} indicates the use of random video during inference.}
    \label{fig:mcdpo_study}
\end{wrapfigure}

\paragraph{Impact of Motion Video Retriever.}
To study the the influence of the video retriever, we replace the Gemini-MVR with two other retrievers: one being the fine-tuned InternVideo~\cite{internvideo} model, and the other being random retrieval from the HcVD.
The results in Table~\ref{tab:ablation} show that random retrieval leads to a notable increase in the FID score, indicating the importance of accurate video priors in generating high-fidelity results. Moreover, replacing the retriever with InternVideo also leads to a rise in FID score, further confirming the effectiveness of Gemini-MVR.
It is important to note that the performance of the retrieval model in RAG is influenced by the generation module, hence we also conduct text-to-video retrieval experiments. 
As shown in Table~\ref{tab:text-to-video}, Gemini-MVR achieves an 8.8\% increase in R@1 for the human-centric video set (pool size is 1990), and a 16.6\% increase in R@1 for the single human-centric video set. 
These results further validate the effectiveness of Gemini-MVR.


\paragraph{Impact of McDPO.}
Table~\ref{tab:ablation} demonstrates that removing McDPO leads to a substantial overall performance drop, as evidenced by the comparison between the \texttt{Gem+Mc} and \texttt{Gem} settings.
These findings indicate that the McDPO trainer effectively mitigates the error propagation issue, a point we further analyze in detail in the discussion section~\ref{discussion_mcdpo}.


\subsection{In-depth Discussion}
\paragraph{The Effect of Retrieval Database Size.}
Figure~\ref{fig:database_size1} illustrates the changes in the FID and MM-Dist metrics as the size of the database increases. 
It can be observed that as the database size increases, both metrics show a decreasing trend, demonstrating VimoRAG’s scalability potential with larger retrieval corpora — a promising property given that wild video datasets can be easily scaled in real-world applications.


\vspace{-2mm}
\paragraph{The Role of McDPO.} 
\label{discussion_mcdpo}

\pgfplotsset{compat=1.7,every axis title/.append style={at={(0.5,-0.45)}, font=\fontsize{10}{1}\selectfont},every axis/.append style={xtick pos=left,ytick pos=left,tickwidth=1.5pt}}
\usetikzlibrary{matrix}
\usepgfplotslibrary{groupplots}
\usetikzlibrary{patterns,backgrounds}

\definecolor{c1}{RGB}{241,169,64}
\definecolor{c2}{RGB}{238,174,238}
\definecolor{c3}{RGB}{53,218,247}
\definecolor{c4}{RGB}{193,236,217}
\definecolor{c5}{RGB}{107,112,092}

\begin{figure}[t!]
\centering
\begin{tikzpicture}
\begin{groupplot}[group style={group name=myplot,group size=2 by 1,horizontal sep=30pt}, height=4cm, width=7cm]

\nextgroupplot[
    smooth,
    ytick={0,0.1,0.12,0.14,0.16,0.18,0.2},
    yticklabels={0,0.1,0.12,0.14,0.16,0.18,0.2},
    ymin=0.1, ymax=0.2,
    ylabel={FID},
    y tick label style={yshift=-0.25em, text height=0ex, font=\small},
    xlabel={Size of Video Database ($\times 10^4$ )},
    y label style={yshift=-0.5em, font=\small, align=center},
    x label style={font=\small},
    axis x line*=bottom,
    axis line style={-},
    axis y line*=left,
    axis line style={-},
    enlargelimits=0.05,
    every node near coord/.append style={black, font=\small, opacity=0.7, yshift=-0.0em, xshift=0.0em},
    legend style={at={(0.5,1.01)}, anchor=south west, draw=none, legend columns=-1, font=\small},
    xticklabels={1.25,2.5,5,10,20,40},
    xtick={1,2,3,4,5,6},
    ymajorgrids,
    xmax=6.2, xmin=0.8,
    x tick label style={yshift=0.05em, align=center, font=\small},
    title style={yshift=-0.9em, font=\small},
]

\addplot[c3, thick, mark=*] coordinates {(1,  0.193) (2, 0.176) (3, 0.174) (4, 0.137) (5, 0.143) (6, 0.124)};

\nextgroupplot[
    smooth,
    ytick={0,3.0,3.1,3.2,3.3,3.4},
    yticklabels={0,3.0,3.1,3.2,3.3,3.4},
    ymin=3.1, ymax=3.4,
    ylabel={MM Dist},
    y tick label style={yshift=-0.25em, text height=0ex, font=\small},
    xlabel={Size of Video Database ($\times 10^4$ )},
    y label style={yshift=-0.5em, font=\small, align=center},
    x label style={font=\small},
    axis x line*=bottom,
    axis line style={-},
    axis y line*=left,
    axis line style={-},
    enlargelimits=0.05,
    every node near coord/.append style={black, font=\small, opacity=0.7, yshift=-0.0em, xshift=0.0em},
    xticklabels={1.25,2.5,5,10,20,40},
    xtick={1,2,3,4,5,6},
    ymajorgrids,
    xmax=6.2, xmin=0.8,
    x tick label style={yshift=0.05em, align=center, font=\small},
    title style={yshift=-0.9em, font=\small},
]

\addplot[c3, thick, mark=*] coordinates {(1, 3.239) (2, 3.274) (3, 3.316) (4, 3.235) (5, 3.270) (6, 3.175)};

\end{groupplot}
\end{tikzpicture}
\caption{As the video retrieval database grows, VimoRAG shows steadily improving performance, demonstrating strong potential for real-world applications.}
\label{fig:database_size1}
\end{figure}
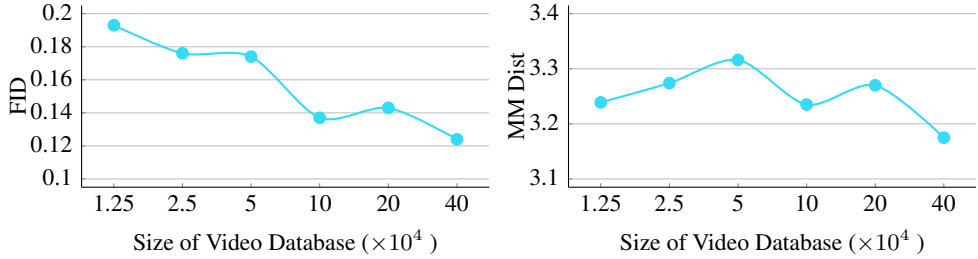

To further analyze the role of McDPO in VimoRAG, we conduct a crossover experiment involving different video priors.
As illustrated in Figure~\ref{fig:mcdpo_study}, the FID score in the \texttt{Mc-R$\infty$} setting is lower than that in the \texttt{NMc-R$\infty$} setting, indicating that McDPO achieves a significantly lower FID score, \textbf{even when random video priors are utilized}. 
This further suggests that McDPO effectively enables the model to disregard non-informative video priors.
Moreover, we observe that McDPO also achieves a significantly lower FID score and higher R-Precision score when provided with a rank-1 video. 
In essence, the model appears to possess the ability to distinguish between informative and non-informative video priors, effectively utilizing relevant information while disregarding noise. 
We hypothesize that this ability arises from the model's implicit alignment with these two perspectives.

\section{Discussion}

\paragraph{Limitations.}
\label{limitation}

A limitation of our work is that VimoRAG is designed for LLMs, which results in longer processing times than those of existing smaller models (motion specialists). 
We evaluate the latency of our framework, with detailed results shown in Table~\ref{tab:efficiency}.
We acknowledge this constraint and intend to explore methods for reducing latency in future works.

\paragraph{Impact Statement.}
\label{impact_statement}

While the capabilities of our framework present significant opportunities for motion generation, they also raise ethical concerns regarding its potential misuse. 
Malicious users could exploit the system to create content that promotes violence or other harmful behaviors, posing risks to societal well-being.
To mitigate these potential impacts, we will implement a strict licensing mechanism upon the release of our method. 
This licensing will govern the academic research and applications of our model, ensuring that its deployment is aligned with ethical standards and responsible use.

\section{Conclusion and Future Work}
\label{future_work}
In this paper, we propose \textbf{\texttt{VimoRAG}}, a novel framework that integrates large-scale in-the-wild video databases to enhance motion generation for motion LLMs. 
We tackle two key challenges—human-centric video retrieval and error propagation—through the proposed Gemini-MVR model and the McDPO training strategy.
Our experiments show that \textbf{\texttt{VimoRAG}} further boosts motion LLMs with substantial performance gains in both OOD and in-domain settings, and its performance steadily improves with larger retrieval corpora, showing strong scalability potential.

In future work, we aim to advance along two directions. First, we will explore which types of LLMs are most suitable as the backbone of \textbf{\texttt{VimoRAG}}, and how to define appropriate metrics for automatically selecting them. While numerous LLMs are available today, our focus in this paper is not on identifying the best-performing LLM, but rather on addressing the core challenges within the RAG system. Second, building upon the success of video-based RAG, we plan to incorporate video, 3D data, and potentially even image data as priors to develop a unified RAG framework and investigating whether this multimodal integration can lead to further performance gains.


\section*{Acknowledgments}
This work was supported in part by Shenzhen Science and Technology Program (JCYJ20241202123503005).
We also acknowledge supports from Guangdong Basic and Applied Basic Research Foundation 2025A1515012281, Nanjing Municipal Science and Technology Bureau 202401035 and University of Macau MYRG-GRG2024-00077-FST-UMDF.

{\small
\bibliographystyle{unsrt}
\bibliography{cite} 
}

\appendix

\clearpage
In the \textbf{appendix}, we present more experimental settings and results (Section~\ref{extended_experiment}), more qualitative results (Section~\ref{extended_qualitative}), details of human-centric video database (Section~\ref{extended_database}), more implementation details of VimoRAG (Section~\ref{extended_vimorag}).

\section{More Experimental Settings and Results}
\label{extended_experiment}
\subsection{Experimental Settings}

In the training configurations for Gemini-MVR, we set the maximum number of video frames to 16. We employ the Adam optimizer with parameters b1=0.9, b2=0.98, epsilon=1e-6, and a weight decay of 0.2 throughout all training stages.
For the action-level retriever, we conduct training for 10 epochs with a batch size of 2048 and a learning rate of 1e-4. 
In the case of the object-level retriever, we train for 5 epochs using a batch size of 128, the learning rate is set to 4e-6 for the CLIP-related modules and 1e-3 for the remaining modules.
Regarding the similarity integrator model, we also train for 5 epochs with a batch size of 128 and a learning rate of 1e-3.
Additional details are available in our code.

For the training configurations of McDPO, during Stage 1, we set the batch size to 64, weight decay to 0.0, and the maximum context length to 4096, employing a bf16 precision format. 
We conduct training for 2 epochs on the HumanML3D training set.
In Stage 2, we train for 1 epoch with a learning rate of 2e-4, weight decay of 0.0, batch size of 8, and $\gamma = 0.1$ as defined in Equation~\ref{equ:dpo}. 
We set the temperature of the LLM to 0.9 across all experiments.
The maximum number of video frames is set to 16 during the generation stage.
Most hyper-parameters are selected through grid search techniques applied to the validation sets.

\pgfplotsset{compat=1.7,every axis title/.append style={at={(0.5,-0.45)}, font=\fontsize{10}{1}\selectfont},every axis/.append style={xtick pos=left,ytick pos=left,tickwidth=1.5pt}}
\usetikzlibrary{matrix}
\usepgfplotslibrary{groupplots}
\usetikzlibrary{patterns,backgrounds}

\definecolor{c1}{RGB}{241,169,64}
\definecolor{c2}{RGB}{238,174,238}
\definecolor{c3}{RGB}{53,218,247}
\definecolor{c4}{RGB}{193,236,217}
\definecolor{c5}{RGB}{250,127,111}

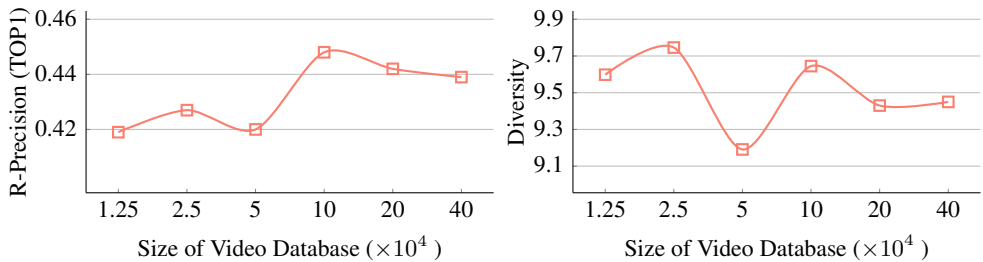
\begin{figure}[htp]
\centering
\begin{tikzpicture}
\begin{groupplot}[group style={group name=myplot,group size=2 by 1,horizontal sep=30pt}, height=4cm, width=7cm]

\nextgroupplot[
    smooth,
    ytick={0,0.42,0.44,0.46,0.48,0.5,0.52},
    yticklabels={0,0.42,0.44,0.46,0.48,0.5,0.52},
    ymin=0.4, ymax=0.46,
    ylabel={R-Precision (TOP1)},
    y tick label style={yshift=-0.25em, text height=0ex, font=\small},
    xlabel={Size of Video Database ($\times 10^4$ )},
    x label style={font=\small},
    y label style={yshift=-0.5em, font=\small, align=center},
    axis x line*=bottom,
    axis line style={-},
    axis y line*=left,
    axis line style={-},
    enlargelimits=0.05,
    every node near coord/.append style={black, font=\small, opacity=0.7, yshift=-0.0em, xshift=0.0em},
    legend style={at={(0.5,1.01)}, anchor=south west, draw=none, legend columns=-1, font=\small},
    xticklabels={1.25,2.5,5,10,20,40},
    xtick={1,2,3,4,5,6},
    ymajorgrids,
    xmax=6.2, xmin=0.8,
    x tick label style={yshift=0.05em, align=center, font=\small},
    title style={yshift=-0.9em, font=\small},
]

\addplot[c5, thick, mark=square] coordinates {(1,  0.419) (2, 0.427) (3, 0.420) (4, 0.448) (5, 0.442) (6, 0.439)};

\nextgroupplot[
    smooth,
    ytick={0,9.1,9.3,9.5,9.7,9.9},
    yticklabels={0,9.1,9.3,9.5,9.7,9.9},
    ymin=9.0, ymax=9.9,
    ylabel={Diversity},
    y tick label style={yshift=-0.25em, text height=0ex, font=\small},
    xlabel={Size of Video Database ($\times 10^4$ )},
    y label style={yshift=-0.5em, font=\small, align=center},
    x label style={font=\small},
    axis x line*=bottom,
    axis line style={-},
    axis y line*=left,
    axis line style={-},
    enlargelimits=0.05,
    every node near coord/.append style={black, font=\small, opacity=0.7, yshift=-0.0em, xshift=0.0em},
    legend style={at={(0.5,1.01)}, anchor=south west, draw=none, legend columns=-1, font=\small},
    xticklabels={1.25,2.5,5,10,20,40},
    xtick={1,2,3,4,5,6},
    ymajorgrids,
    xmax=6.2, xmin=0.8,
    x tick label style={yshift=0.05em, align=center, font=\small},
    title style={yshift=-0.9em, font=\small},
]

\addplot[c5, thick, mark=square] coordinates {(1, 9.598) (2, 9.746) (3, 9.191) (4, 9.645) (5, 9.430) (6, 9.450)};

\end{groupplot}
\end{tikzpicture}
\caption{Variation of \textit{R-Precision} (TOP1) and \textit{Diversity} metrics as a function of video database size.
A larger database, exceeding 100,000 entries, enhances R-precision. There appears to be no evident correlation between the diversity metric and the size of the database.}
\label{fig:database_size2}
\end{figure}

\begin{table*}[h]
    \centering
    \fontsize{9}{10}\selectfont
    \setlength{\tabcolsep}{3.5mm}
    \caption{Performance of \textbf{various LoRA parameters} in McDPO on the HumanML3D validation set. The best results are highlighted \textbf{in bold} for each setting.}
    \begin{tabular}{cc|cccccc}
         \toprule
             \multirow{2}{*}{$Rank$} & \multirow{2}{*}{$\alpha$} & \multirow{2}{*}{\textbf{FID} $\downarrow$} & \multirow{2}{*}{\textbf{MultiModal Dist} $\downarrow$} & \multicolumn{3}{c}{\textbf{R-Precision} $\uparrow$} & \multirow{2}{*}{\textbf{Diversity} $\uparrow$} \\
             \cmidrule{5-7}
            & & & & Top-1 & Top-2 & Top-3 & \\
        \midrule
            \rowcolor{lightgray} \multicolumn{8}{l}{\bf\em $\bullet$ $\alpha/rank=2$}\\
            8 & 16 &  0.417   &  3.216  & 0.437 & 0.625 &  0.745  & 9.636 \\
            16 & 32 & 0.249 & 3.146 & 0.426 & 0.642 & 0.764 & \textbf{9.936} \\
            32 & 64 & 0.221  & 3.109 & 0.449 & \textbf{0.670} & 0.759  & 9.718 \\
            64 & 128 & 0.192  & 3.089 & \textbf{0.454} & 0.664 & 0.767  & 9.560 \\
            128 & 256 & \textbf{0.179}  & \textbf{3.046} & 0.447 & 0.667 & \textbf{0.772}  & 9.567 \\
        \midrule
            \rowcolor{lightgray} \multicolumn{8}{l}{\bf\em $\bullet$ $rank=128$}\\
            128 & 64 &  0.235  & 3.110   &  0.452  &  0.656  & 0.770   &  9.591  \\
            128 & 128 & \textbf{0.156}  & 3.114    &  0.445  &  0.637   &  0.754  & \textbf{9.728} \\
            128 & 256 & 0.179  & \textbf{3.046} & \textbf{0.447} & \textbf{0.667} & \textbf{0.772}  & 9.567 \\
        \midrule
            \rowcolor{lightgray} \multicolumn{8}{l}{\bf\em $\bullet$ $\alpha=256$}\\
            32 & 256 & 0.270 &  3.184  & 0.437  &  0.638  &  0.742 & 9.294 \\
            64 & 256 & \textbf{0.152}    &  3.168   &  0.443  &  0.654   &   0.761  &  9.486   \\
            128 & 256 & 0.179  & \textbf{3.046} & \textbf{0.447} & \textbf{0.667} & \textbf{0.772}  & \textbf{9.567} \\
        \bottomrule
    \end{tabular}
    \label{tab:lora}
    \vspace{-0.25cm}
\end{table*}

\begin{table}[!t]
\fontsize{8}{10}\selectfont
\setlength{\tabcolsep}{3mm}
\caption{
Analysis of average latency per instance during the inference phase (in seconds).
 \textit{Retrieval} refers to the process of transforming text into video, \textit{Generation} indicates the phase of generating motion tokens, and \textit{Decoding} pertains to the conversion of tokens into features using VQ-VAE. It is evident that the retrieval phase does not represent a bottleneck within the entire pipeline, rather, the generation process of LLM constitutes the limiting factor. Notably, the parameter size of the LLM utilized in VimoRAG is 3.8 billion.
}
\begin{center}
 \setlength{\tabcolsep}{2.0mm}
\begin{tabular}{lcccccccccc}

\toprule

&\bf Retrieval Time (s)&\bf Generation Time (s)&\bf Decoding Time (s)&\bf Total Time (s)&\bf Tokens/s\\
\midrule
MotionGPT-13B &0 & 14.89 & 0.12 & 15.01 &13.44\\
VimoRAG &0.48 & 7.02 & 0.12 & 7.62 & 27.92\\
\bottomrule
\end{tabular}
\label{tab:efficiency}
\end{center}
\end{table}

\begin{table*}[!t]
    \fontsize{8.0}{10.0}\selectfont
    \setlength{\tabcolsep}{0.8mm}
        \centering
        \caption{Complete results with confidence intervals on the HumanML3D test set.}
        \label{tab:t2m_full}
            \begin{tabular}{llllllll}%
                \toprule
                \multirow{2}{*}{\textbf{Model}} & \multirow{2}{*}{\textbf{Backbone}}& \multirow{2}{*}{\textbf{FID} $\downarrow$}& \multicolumn{3}{c}{\textbf{R-Precision} $\uparrow$}  & \multirow{2}{*}{\textbf{MM Dist} $\downarrow$} & \multirow{2}{*}{\textbf{Diversity} $\uparrow$} \\
                \cmidrule{4-6}
                & & &Top 1 &  Top 2 & Top 3 & \\
                \midrule
                \rowcolor{lightgray} \multicolumn{8}{l}{\bf\em $\bullet$ Motion Specialists}\\
                MoMask~\cite{momask}    &--& 0.048$^{\pm.004}$          & 0.519$^{\pm.005}$          & 0.715$^{\pm.005}$          & 0.809$^{\pm.004}$          & 2.955$^{\pm.011}$          & 9.632$^{\pm.094}$                           \\
                T2M-GPT~\cite{t2m-gpt}    &--               & 0.112$^{\pm.006}$ & 0.489$^{\pm.006}$ & 0.679$^{\pm.006}$ & 0.774$^{\pm.004}$          & 3.125$^{\pm.015}$ & 9.691$^{\pm.062}$                     \\
                MDM~\cite{mdm}             &--         & 0.454$^{\pm.012}$          & 0.419$^{\pm.004}$          & 0.606$^{\pm.004}$          & 0.712$^{\pm.004}$          & 3.636$^{\pm.015}$          & 9.449$^{\pm.136}$            \\
                MotionDiffuse~\cite{motion_diffuse}  &                    --& 0.672$^{\pm.025}$          & 0.492$^{\pm.004}$          & 0.685$^{\pm.003}$          & 0.784$^{\pm.003}$          & 3.085$^{\pm.134}$          & 9.499$^{\pm.184}$  \\
                MLD~\cite{mld}   &       --& 0.425$^{\pm.145}$          & 0.468$^{\pm.005}$          & 0.656$^{\pm.003}$          & 0.759$^{\pm.004}$ & 3.266$^{\pm.019}$          & 9.698$^{\pm.094}$           \\
                ReMoDiffuse~\cite{remodiffuse}   &--       & 0.125$^{\pm.142}$          & 0.493$^{\pm.004}$          & 0.676$^{\pm.003}$          & 0.775$^{\pm.003}$ & 3.047$^{\pm.007}$          & 9.211$^{\pm.129}$    \\
                LMM*~\cite{lmm}      & --                       & 0.040$^{\pm.002}$ & 0.525$^{\pm.002}$ & 0.719$^{\pm.002}$ & 0.811$^{\pm.002}$ & 2.943$^{\pm.012}$ & 9.814$^{\pm.076}$             \\
                MotionLab*~\cite{motionlab} &--                             & 0.167$^{\pm-}$ & --$^{}$ & --$^{}$ & 0.810$^{\pm-}$ & 2.912$^{\pm-}$ & 9.593$^{\pm-}$             \\
                MotionLCM*~\cite{motionlcm} &--                             & 0.304$^{\pm.012}$ & 0.502$^{\pm.003}$ & 0.698$^{\pm.002}$ & 0.798$^{\pm.002}$ & 3.012$^{\pm.007}$ & 9.607$^{\pm.006}$             \\
                MotionCLR*~\cite{motionclr} &--                             & 0.269$^{\pm.001}$ & 0.544$^{\pm.001}$ & 0.732$^{\pm.001}$ & 0.831$^{\pm.002}$ & 2.806$^{\pm.014}$ & --$^{}$             \\
                MotionGPT*~\cite{motiongpt_nips} &--                             & 0.232$^{\pm.008}$ & 0.492$^{\pm.003}$ & 0.681$^{\pm.003}$ & 0.778$^{\pm.002}$ & 3.096$^{\pm.008}$ & 9.528$^{\pm.071}$             \\
                BiPO*~\cite{bipo} &--                             & \textbf{0.030}$^{\pm.002}$ & 0.523$^{\pm.003}$ & 0.714$^{\pm.002}$ & 0.809$^{\pm.002}$ & 2.880$^{\pm.009}$ & 9.556$^{\pm.076}$             \\
                StableMoFusion*~\cite{stable_mofusion} &--                             & 0.098$^{\pm.003}$ & 0.553$^{\pm.003}$ & 0.748$^{\pm.002}$ & 0.841$^{\pm.002}$ & --$^{}$ & 9.748$^{\pm.092}$             \\
                MoGenTS*~\cite{mogents} &--                             & 0.033$^{\pm.001}$ & 0.529$^{\pm.003}$ & 0.719$^{\pm.002}$ & 0.812$^{\pm.002}$ & 2.867$^{\pm.006}$ & 9.570$^{\pm.077}$             \\
                LAMP*~\cite{lamp} &--                             & 0.032$^{\pm.002}$ & \textbf{0.557}$^{\pm.003}$ & \textbf{0.751}$^{\pm.002}$ & \textbf{0.843}$^{\pm.001}$ & \textbf{2.759}$^{\pm.007}$ & 9.571$^{\pm.069}$             \\
                \midrule
                \rowcolor{lightgray} \multicolumn{8}{l}{\bf\em $\bullet$ Motion LLMs}\\
                MotionGPT-2*~\cite{motiongpt2} &Llama3-8B                             & 0.191$^{\pm.004}$ & 0.496$^{\pm.002}$ & 0.691$^{\pm.003}$ & 0.782$^{\pm.004}$ & 3.080$^{\pm.013}$ & 9.860$^{\pm.026}$             \\
    
                MotionLLM*~\cite{motionagent} &GPT4+Gemma-2B                             & 0.230$^{\pm.009}$ & 0.515$^{\pm.004}$ & --$^{}$ & 0.801$^{\pm.004}$ & 2.967$^{\pm.020}$ & \textbf{9.908}$^{\pm.102}$             \\
                \textit{Wang et al.}*~\cite{largemotionmodel} &Llama2-13B                             & 0.166$^{\pm-}$ & 0.519$^{\pm-}$ & --$^{}$ & 0.803$^{\pm-}$ & 2.964$^{\pm-}$ & --$^{}$             \\
                ScaMo*~\cite{scamo} &codesize 512-3B                             & 0.617$^{\pm-}$ & 0.443$^{\pm-}$ & 0.627$^{\pm-}$ & 0.734$^{\pm-}$ & 3.340$^{\pm-}$ & 9.217$^{\pm-}$             \\
                AvatarGPT*~\cite{avatargpt} &Llama-13B                             & 0.567$^{\pm-}$ & 0.389$^{\pm-}$ & 0.539$^{\pm-}$ & 0.623$^{\pm-}$ & --$^{}$ & 9.489$^{\pm-}$             \\
                MotionGPT*~\cite{motiongpt_aaai}  &Llama-13B      & 0.567$^{\pm-}$          & --$^{}$          & --$^{}$          & --$^{}$          & 3.775$^{\pm-}$          & 9.006$^{\pm-}$   \\
                \midrule
                MotionGPT~\cite{motiongpt_aaai}    & Phi3-3.8B    & 0.501$^{\pm.005}$          & 0.396$^{\pm.002}$          & 0.575$^{\pm.005}$          & 0.673$^{\pm.004}$          & 3.724$^{\pm.012}$          & 9.475$^{\pm.110}$ \\
                VimoRAG~(Ours)    & Phi3-3.8B    & 0.131$^{\pm.007}$
              & 0.452$^{\pm.002}$          & 0.655$^{\pm.006}$          & 0.764$^{\pm.005}$          & 3.146$^{\pm.011}$          & 9.424$^{\pm.149}$ \\
                \bottomrule
            \end{tabular}%
    \vspace{-3mm}
    \end{table*}

\subsection{Metrics}
For motion generation evaluation metrics, we adopt several widely recognized measures: \textit{Frechet Inception Distance} (FID), 
which quantifies generation fidelity by measuring the distributional distance between the generated motions and reference motions in feature space.
\textit{R-recall}, \textit{MultiModal Distance} (MM Dist), 
which evaluate the semantic consistency between text and motions.
\textit{Diversity}.
which assesses the diversity of the generated motions corresponding to a given textual input.
For text-to-video retrieval metrics, we adopt the widely used metrics retrieval \textit{recall} (R@1, R@5, R@10 are adopted in this paper), \textit{Median Rank} (MdR) and \textit{Mean Rank} (MnR).
Recall measures the proportion of relevant results returned by the retrieval model within the top-k results.
Median Rank represents the middle value of the ranks at which the correct results appear in the retrieval list.
Mean Rank calculates the average position of the correct results in the retrieval lists.

\subsection{More Details of Datasets}

IDEA400 represents a high-quality subset of Motion-X~\cite{motionx}, consisting of a large-scale whole-body motion dataset composed of 12.5K clips and 2.6M frames. 
This dataset encompasses a diverse array of gestures and detailed pose descriptions. 
Unlike existing works~\cite{plan_posture}, which meticulously select a test set with text descriptions similar to those found in HumanML3D, we adopt a different approach by randomly sampling 10\% of the clips without any specific selection criteria. 
We argue that this methodology more closely resembles an out-of-distribution (OOD) scenario, wherein the test set features a distribution distinctly different from that of the HumanML3D training set.

HumanML3D~\cite{humanml3d} constitutes the largest dataset available, providing text descriptions alongside body-only motions. 
It includes a total of 14,616 motion clips and 44,970 text descriptions, with 5,371 unique words present across these descriptions. 
The dataset is partitioned into a training set (80\%), a validation set (5\%), and a test set (15\%).

Regarding pose representation, we adhere to the same specifications as outlined in HumanML3D~\cite{humanml3d}. 
It is important to note that the number of joints ($J$) is set at 22 for both HumanML3D and IDEA400.
We process the motion features for IDEA400 using the same settings as HumanML3D, resulting in a total dimension of 263. 
These features include root angular velocity, root linear velocities, root height, local joint positions (velocities), and 6D rotations.

\begin{figure}[!t]
    \centering
    \includegraphics[width=0.98\linewidth]{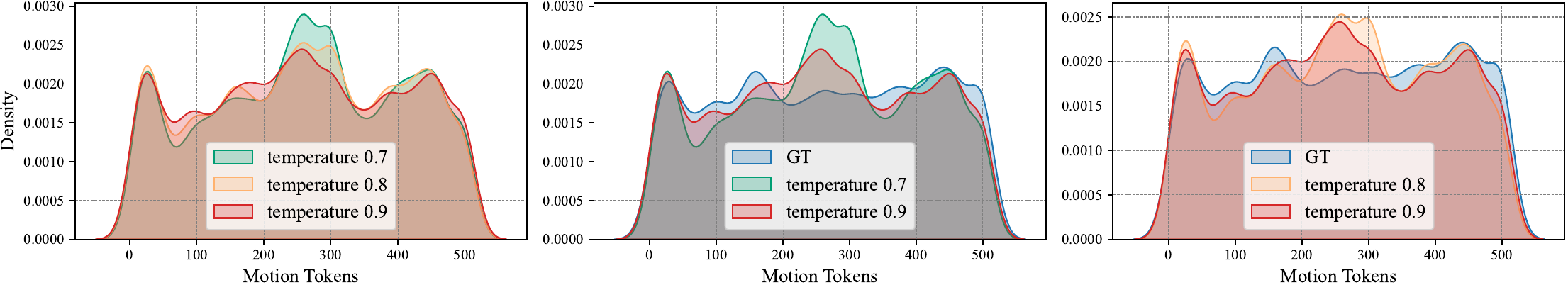}
    \caption{The impact of varying temperature values on the distribution of generated motion tokens. Notably, when the temperature is set to 0.9, the distribution of the generated motion tokens closely resembles the ground truth (GT) distribution.}
    \label{fig:distribution_temperature}
\end{figure}

\begin{figure}[!t]
    \centering
    \includegraphics[width=0.98\linewidth]{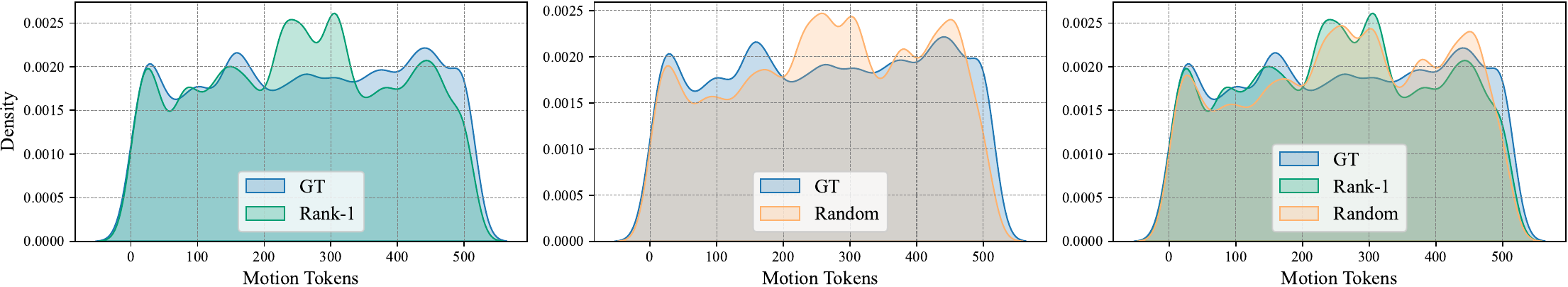}
    \caption{The influence of different retrieval conditions on the distribution of generated motion tokens is depicted.}
    \label{fig:distribution_retrieval}
\end{figure}

\subsection{More Experimental Results}
\paragraph{The Effect of Video Database Size.}

To investigate the impact of video database size on performance, we conduct experiments at six different scales using the validation set from the HumanML3D dataset. Figure~\ref{fig:database_size1} illustrates the changes in the \textit{Frechet Inception Distance} (FID) and \textit{Multi-Modal Distance} (MM-Dist) metrics as the size of the database increases,
as discussed in Section~\ref{sec:main_results}.

We also report the changes in R-Precision and diversity metrics under the same experimental settings. 
As illustrated in Figure~\ref{fig:database_size2}, we observe that a larger database contributes to an improvement in R-Precision, particularly when the database size exceeds 100,000 entries, highlighting the advantages of utilizing large-scale video retrieval repositories. 
Interestingly, we find that the diversity metric does not increase in tandem with the growth of the video database size. We hypothesize that when the retrieved videos lack informativeness, the noise introduced enhances the diversity of the model's outputs. 
Consequently, smaller video databases can still yield substantial diversity.

Through this experiment, we demonstrate the significant potential of retrieval-enhanced methods based on large-scale video databases. 
However, resource constraints limit us to conduct further experiments. 
We hope to explore the effects of even larger databases on performance in future work.

\paragraph{The Analysis of Latency.}
We analyze the latency of the VimoRAG framework, as depicted in Table~\ref{tab:efficiency}, where we report the average time taken per instance during the inference phase. 
The entire pipeline is divided into three stages: the retrieval phase (conducted by the text-to-video model), the generation phase (performed by the LLM), and the decoding phase (executed by the VQ-VAE).
The results indicate that the generation phase exhibits the highest latency, accounting for 92\% of the total processing time. 
This is primarily due to the need for the LLM to load a substantial number of parameters (3.8 billion in VimoRAG). 
In comparison, the retrieval and decoding phases account for only 6\% and 2\% of the total time, respectively. 
This finding highlights that, despite the generation of high-fidelity motion leveraging the world knowledge of the LLM, it remains the latency bottleneck in the overall generation framework. 
Consequently, this drives us to explore more efficient generative models in our future work.
To illustrate the impact of different LLM sizes on latency, we also test the latency of the MotionGPT-13B model~\cite{motiongpt_aaai} on the same computational hardware. 
As shown in Table~\ref{tab:efficiency}, the total time for the MotionGPT-13B is nearly twice that of VimoRAG.

\begin{figure}[!t]
    \centering
    \includegraphics[width=0.87\linewidth]{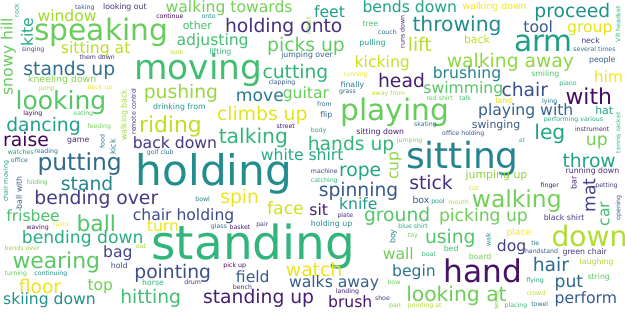}
    \caption{Words and phrases that frequently appear in text descriptions in HcVD database.}
    \label{fig:word_cloud}
\end{figure}

\paragraph{Hyper-parameters of LoRA in McDPO.}
In order to investigate the impact of key hyper-parameters on the performance of McDPO training, we conduct comparative experiments focusing on the LoRA fine-tuning parameters $rank$ and $\alpha$. 
As illustrated in Table 1, varying $rank$ and $\alpha$ significantly influence the outcomes, a phenomenon that aligns with previous findings in the works~\cite{motiongpt_aaai,motiongpt2}. 
We also observe that when maintaining a constant ratio between \( rank \) and \( \alpha \), both the FID and MultiModal Dist metrics gradually decrease as \( rank \) and \( \alpha \) increase. The increase in \( rank \) and \( \alpha \) corresponds to a greater number of parameters available for optimization within the model, thereby enhancing its capability to fit the dataset more effectively. 
However, we note a corresponding decline in diversity, indicating that a larger parameter scale adversely affects the model's ability to generate diverse outputs.

\paragraph{The Impact of the Temperature of LLM.}
As shown in the Figure~\ref{fig:distribution_temperature}, we visualize the impact of different temperatures on the distribution of generated motion tokens. We examine the effects of three commonly used temperatures (0.7, 0.8, and 0.9) on the results. Firstly, it is evident that the output distributions generated by different temperatures exhibit differences, indicating that the model is sensitive to this hyperparameter. 
Additionally, we observe that the distribution at a temperature of 0.9 is closest to the ground truth distribution. 
Generally, as the temperature parameter increases, the output diversity of the model also rises. 
We believe that a larger temperature in this study aids in enhancing the model's generalization capability.

\paragraph{Analysis of Distribution Resulting from the Retrieval Process.}
As shown in Figure~\ref{fig:distribution_retrieval}, we visualize the distribution of generated motion tokens under two conditions: one utilizing rank-1 videos during inference and the other using random videos.
The figure demonstrates that the distributions of motion tokens obtained from random videos and rank-1 videos are generally comparable on a macroscopic level. 
This indicates that even when the retrieval system fails, the distribution of our model's generated results remains relatively close to the distributions obtained from rank-1 retrieval and the ground truth distribution from a macroscopic perspective. 
This demonstrates the model's strong fault tolerance and robustness.
Notably, when the motion token index is less than 200, the generated distribution under the rank-1 condition closely aligns with the ground truth distribution, indicating the effectiveness of this retrieval strategy in capturing motion characteristics.

\section{More Qualitative Results}
\label{extended_qualitative}
\begin{figure}[!t]
    \centering
    \includegraphics[width=1.0\linewidth]{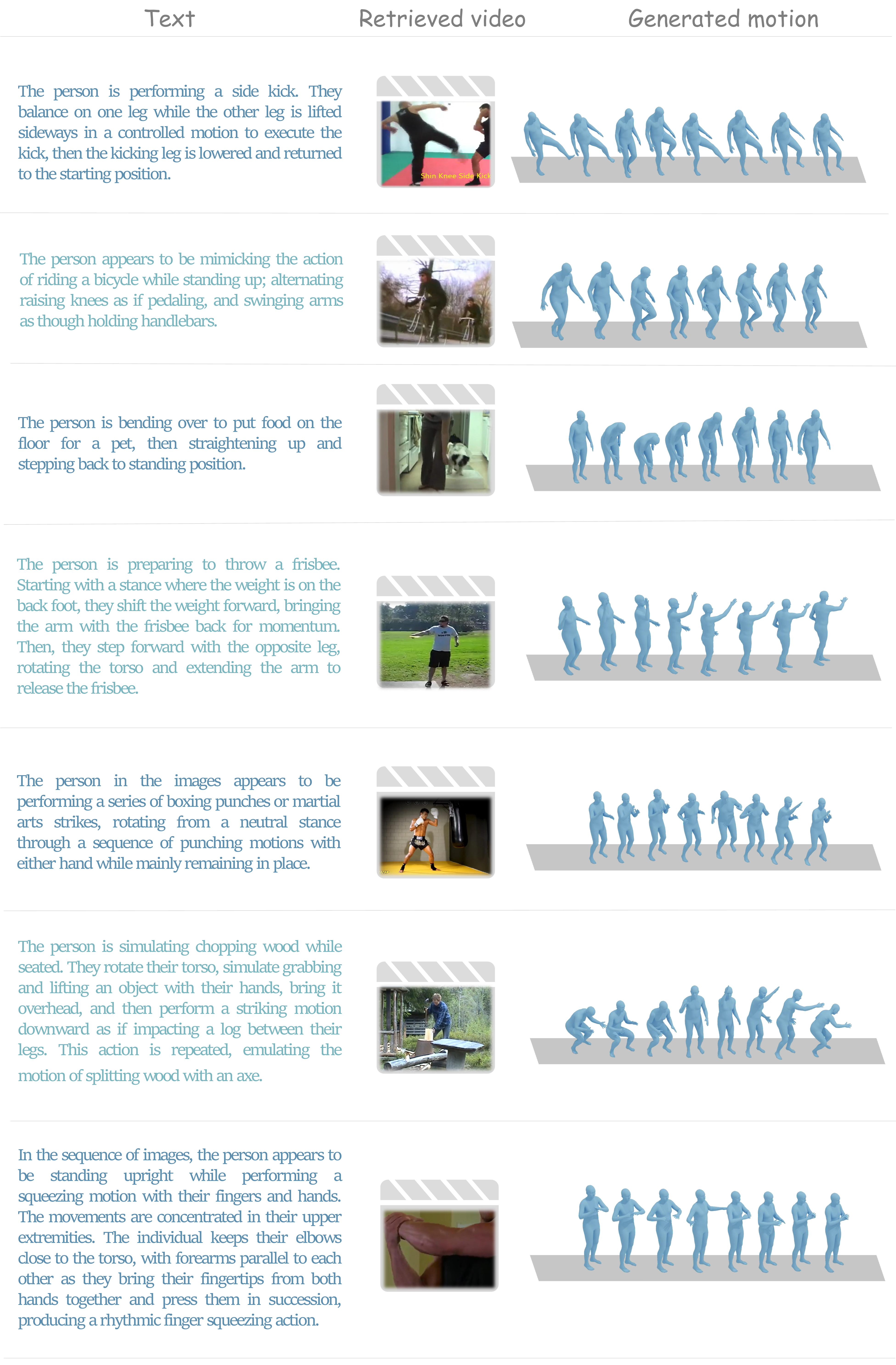}
    \caption{
Additional visualization results on the IDEA400 dataset (\textbf{Part} \uppercase\expandafter{\romannumeral1}).}
    \label{fig:appendix1}
\end{figure}

\begin{figure}[!t]
    \centering
    \includegraphics[width=1.0\linewidth]{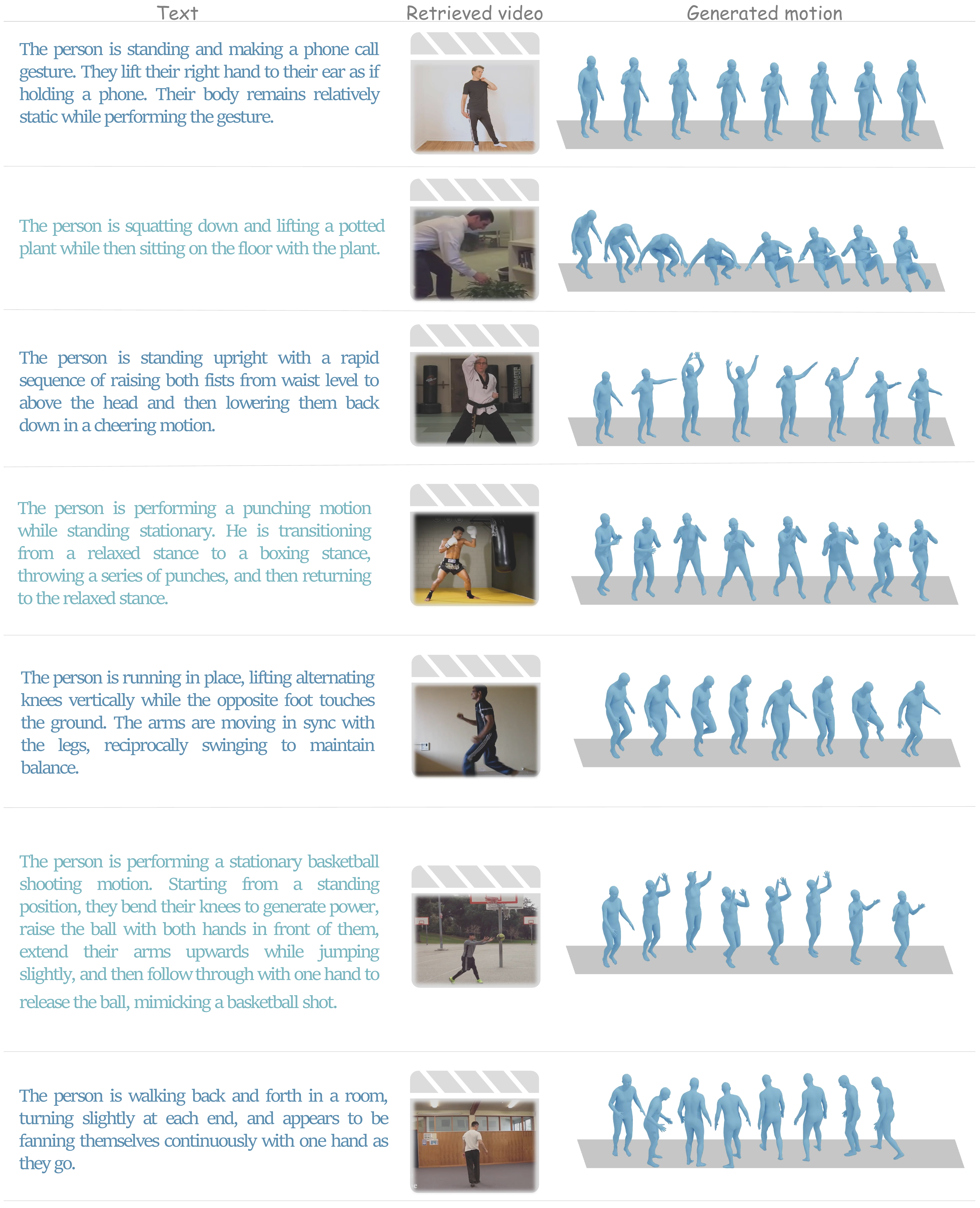}
    \caption{Additional visualization results on the IDEA400 dataset (\textbf{Part} \uppercase\expandafter{\romannumeral2}).}
    \label{fig:appendix2}
\end{figure}

To comprehensively evaluate the generation performance of VimoRAG in out-of-distribution (OOD) scenarios, we present more visualization results on the IDEA400 dataset in Figures~\ref{fig:appendix1} and~\ref{fig:appendix2}. 
It is noteworthy that the model utilized for generation is trained solely on the HumanML3D training set.
We also present the retrieved videos on the right side of each case. 
The cases depicted in Figure~\ref{fig:idea400} are further illustrated in these two figures. 
In contrast, we provide full-text descriptions in these figures. 
It is evident that the generated motions align with the text descriptions, aided by the retrieved videos. Taking a complex description, \textit{``The person is simulating chopping wood while seated. They rotate their torso, simulate grabbing and lifting an object with their hands, bring it overhead, and then perform a striking motion downward as if impacting a log between their legs. This action is repeated, emulating the motion of splitting wood with an axe.''} in Figure~\ref{fig:appendix1}, as an example, VimoRAG is capable of generating such uncommon actions and seamless transitions between movements.
As a supplement, we have showcased some of the original videos in our anonymous repository.

\section{Details of Human-centric Video Database}
\label{extended_database}

To train our retrieval models, we annotate a text description for each video using the widely utilized LMM, Qwen2-VL-7B-Instruct. In total, we synthesize 425,988 captions. 
It is important to note that we use these text captions exclusively during the training phase of Gemini-MVR and do not employ them in any retrieval processes within VimoRAG. 
This means that VimoRAG also performs effectively with another large video database during the inference stage. As illustrated in Figure~\ref{fig:word_cloud}, the word cloud presents the diverse types of actions included in the HcVD database. 
This richness and diversity also elucidate why VimoRAG achieves exceptional performance from a different perspective.
We present more details of the resources that are used for the construction of the HcVD in Table~\ref{tab:resources}.
Qwen2-VL supports dynamic frame selection (set FPS as 2.0).
The prompt we used for data synthesis is ``Please describe the person's actions in the video using a single sentence that contains a series of verbs.''.

\begin{table}[!t]

\fontsize{9}{8.0}\selectfont
    \centering
    \caption{Details of the resources utilized in the construction of the HcVD database.}
    \renewcommand{\arraystretch}{1.3} 
    \begin{tabular}{@{}lcl@{}}
        \toprule
        \textbf{Dataset} & \textbf{Number of Videos} & \textbf{Tasks} \\ \midrule
        UCF101~\cite{ucf-101}& 13,320& Action Recognition\\ 
        NTU RGB+D~\cite{nturgb}& 114,480 & Action Recognition\\
        ASLAN~\cite{aslan}& 3,697& Action Similarity Labeling \\ 
        HMDB51~\cite{hmdb51}& 6,849 & Human Motion Recognition\\ 
        Kinetics-400~\cite{kinetics-400}& 306,245& Human Action Classification\\ 
        PennAction~\cite{penn_action}& 2,326& Action Classification, Action Detection\\ 
        MotionX~\cite{motionx}& 32,500& Human Mesh Recovery, Human Mesh Generation\\  \bottomrule
    \end{tabular}
    \label{tab:resources}
\end{table}

\section{More Implementation Details of VimoRAG}
\label{extended_vimorag}
To enhance the reproducibility of VimoRAG, this section presents additional model details, which are also available in our code repository.
\paragraph{More Details of the Gemini-MVR.}

The temporal encoder consists of 4 transformer layers, each featuring 12 attention heads and a width of 768. We utilize learnable position embeddings with a context length of 77 in the action encoder. 
For the keypoints encoder, a projection layer is placed atop MotionBERT~\cite{motionbert}, with the input and output channels of the projection layer set at 8704 and 768, respectively. 
To derive the final representation of the keypoints, we implement mean pooling over all the encoded frames, in accordance with existing works~\cite{clip4clip,internvideo}. 
The similarity integrator is implemented as a linear transformation, with an input channel of 768 and an output channel of 2.
The $\mathcal{L}_{a2p}$ is defined as follows:
\begin{equation}
    \mathcal{L}_{a2p} = - \frac{1}{B} \sum_{i}^{B} \log \frac{\exp(s(\textbf{a}_{i},\textbf{p}_{i}))}{ {\textstyle \sum_{j=1}^{B}\exp(s(\textbf{a}_{i},\textbf{p}_{j}))} }  \,.
\end{equation}

\paragraph{More Details of the Generation Model.}

In our framework, we employ the VFM InternVideo2~\cite{internvideo2} as the video encoder and CLIP-large~\cite{clip} as the image encoder following \textit{Maaz et al.}~\cite{videogpt+}.
Specifically, we adopt the \texttt{InternVideo2-Stage2\_1B-224p-f4} variant of InternVideo2 and the \texttt{CLIP-ViT-L/14-336} model.
The visual projector is a two-layer MLP with a hidden size of 1024, where a GELU~\cite{gelu} activation function is applied after the first linear layer.
For the LoRA parameters, we configure the rank to 128 and set alpha to 256.
For the model details and training configurations of VQ-VAE, we adopt the same settings as those used in existing works~\cite{t2m-gpt,motiongpt_aaai}. Specifically, the codebook size is set to 512x512.
The temporal downsampling rate is set to 3 in the encoder of VQ-VAE.

\paragraph{More Details of the McDPO Dataset.}
Firstly, we use the $\pi_{ref}$ model obtained from Stage 1 to sample \(k\) times (where \(k=3\)) on a random 25\% subset of the training set. 
The reason for extracting this subset is to reduce inference costs (which is a similar approach used by \textit{Zhang et al.}~\cite{llava-hound}). 
For each input sample, we obtain \(k\) different outputs as candidate samples. 
To identify positive and negative samples from these multiple outputs, we utilize Equation ~\ref{equ:reward} to calculate the reward scores and select the cases with the highest and lowest scores as the positive and negative samples, respectively.
We configure $w_{\ell} = 0.9$ and $w_{d} = 0.1$ in Equation~\ref{equ:reward} in our experiments.

\end{document}